%% file: main.tex
\documentclass[journal]{IEEEtran}
\usepackage{amsmath,amsfonts}
\usepackage{algorithmic}
\usepackage[ruled,linesnumbered]{algorithm2e}
\usepackage{bm}
\usepackage[caption=false,font=normalsize,labelfont=sf,textfont=sf]{subfig}
\usepackage{textcomp}
\usepackage{stfloats}
\usepackage{url}
\usepackage{verbatim}
\usepackage{graphicx}
\usepackage{cite}
\usepackage{tabularx,booktabs}
\usepackage{multirow}
\usepackage{multicol}
\usepackage{colortbl} 
\usepackage{xcolor}
\usepackage{hyperref}
\hypersetup{
	colorlinks=true,
	linkcolor=red,
	anchorcolor=red,
	citecolor=green}
\usepackage{amssymb}
\usepackage{bbding}

\newcolumntype{L}[1]{>{\raggedright\arraybackslash}p{#1}}
\newcolumntype{C}[1]{>{\centering\arraybackslash}p{#1}}
\newcolumntype{R}[1]{>{\raggedleft\arraybackslash}p{#1}}

\newcommand{\tx}{\textcolor{black}}

\makeatletter
\newcommand{\removelatexerror}{\let\@latex@error\@gobble}
\renewcommand{\maketag@@@}[1]{\hbox{\m@th\normalsize\normalfont#1}}%
\makeatother
\usepackage{makeidx}
\makeindex

\begin{document}
	\title{DOMAIN: milDly cOnservative Model-bAsed offlINe reinforcement learning}
	\author{Xiao-Yin Liu,~\IEEEmembership{}  Xiao-Hu Zhou*,~\IEEEmembership{}  Mei-Jiang Gui,~\IEEEmembership{} Guo-Tao Li,~\IEEEmembership{} Xiao-Liang Xie,~\IEEEmembership{}  Shi-Qi Liu,~\IEEEmembership{} Shuang-Yi Wang,~\IEEEmembership{} Qi-Chao Zhang, ~\IEEEmembership{} Biao Luo,~\IEEEmembership{} Zeng-Guang Hou*,~\IEEEmembership{Fellow, IEEE}
  
	\thanks{ This work was supported in part by the National Key Research and Development Program of China under Grant 2023YFC2415100, in part by the National Natural Science Foundation of China under Grant 62222316, Grant 62373351, Grant 82327801, Grant 62073325, Grant 62303463, in part by the Chinese Academy of Sciences Project for Young Scientists in Basic Research under Grant No. YSBR-104, in part by the Beijing Natural Science Foundation under Grant F252068, Grant 4254107, in part by China Postdoctoral Science Foundation under Grant 2024M763535 and in part by CAMS Innovation Fund for Medical Sciences (ClFMS) under Grant 2023-I2M-C\&T-B-017 (*Corresponding authors: Xiao-Hu Zhou and Zeng-Guang Hou).}
	\thanks{Xiao-Yin Liu,  Xiao-Hu Zhou, Mei-Jiang Gui, Guo-Tao Li, Xiao-Liang Xie, Shi-Qi Liu, Shuang-Yi Wang and Qi-Chao Zhang are with the State Key Laboratory of Multimodal Artificial Intelligence Systems, Institute of Automation, Chinese Academy of Sciences, Beijing 100190, China, and also with the School of Artificial Intelligence, University of Chinese Academy of Sciences, Beijing 100049, China. (e-mail: xiaohu.zhou@ia.ac.cn).}
    		
    \thanks{Biao Luo is with the School of Automation, Central South University, Changsha 410083, China (e-mail: biao.luo@hotmail.com).}
        
    \thanks{Zeng-Guang Hou is with the State Key Laboratory of Multimodal Artificial Intelligence Systems, Institute of Automation, Chinese Academy of Sciences, Beijing 100190, China, also with the School of Artificial Intelligence, University of Chinese Academy of Sciences, Beijing 100049, China, and also with CASIA-MUST Joint Laboratory of Intelligence Science and Technology, Institute of Systems Engineering, Macau University of Science and Technology, Macao, China. (e-mail: zengguang.hou@ia.ac.cn).}
	}
	\markboth{}%
	{Shell \MakeLowercase{\textit{et al.}}: A Sample Article Using IEEEtran.cls for IEEE Journals}
	\maketitle
	
	\begin{abstract}
		Model-based reinforcement learning (RL), which learns an environment model from the offline dataset and generates more out-of-distribution model data, has become an effective approach to the problem of distribution shift in offline RL. Due to the gap between the learned and actual environment, conservatism should be incorporated into the algorithm to balance accurate offline data and imprecise model data. The conservatism of current algorithms mostly relies on model uncertainty estimation. \tx{However, uncertainty estimation is unreliable and leads to poor performance in certain scenarios, and the previous methods ignore differences between the model data, which brings great conservatism.} \tx{To address the above issues, this paper proposes a milDly cOnservative Model-bAsed offlINe RL algorithm (DOMAIN) without estimating model uncertainty, and designs the adaptive sampling distribution of model samples, which can adaptively adjust the model data penalty.} In this paper, we theoretically demonstrate that the \emph{Q} value learned by the DOMAIN outside the region is a lower bound of the true \emph{Q} value, the DOMAIN is less conservative than previous model-based offline RL algorithms, and has the guarantee of safety policy improvement. \tx{The results of extensive experiments show that DOMAIN outperforms prior RL algorithms and the average performance has improved by $1.8\%$ on the D4RL benchmark.} 
	\end{abstract}
	\input{tex/01-Introduction.tex}
        \input{tex/09-Related}
	\input{tex/02-Preliminaries.tex}

	\input{tex/03-Method.tex}
	\input{tex/04-Experiment.tex}
	\input{tex/05-Conclusion.tex}

\input{tex/08-Reference.tex}
        \input{tex/06-Appendix.tex}

\end{document}

%% file: tex/01-Introduction.tex
\section{Introduction}
\IEEEPARstart{R}{einforcement} learning (RL) has been widely applied in various fields, including robot manipulation control \cite{ref1,bbbbb_4}, medical robotics control \cite{ref40,bbbbb_1}, underwater robotics control \cite{bbbbb_8,ref3}, and resource scheduling \cite{ref4}. \tx{RL learns the optimal policy through the agent interacting with the environment constantly, that is, trial-and-error \cite{ref36}. However, agents directly interacting with the environment may bring low sampling efficiency and high interaction cost \cite{ref38}, which makes it impractical in certain real-world scenarios. Therefore, offline RL has been widely studied in recent years.}

\tx{Offline RL, which learns the policy froma  static dataset without interaction with the environment, has become an effective way to address the above problem. However, since limited coverage of the collected dataset, offline RL suffers from distribution shift,} which is caused by the difference between the policy in the offline dataset and those learned by the agent \cite{r_2}. The selection of out-of-distribution (OOD) data introduces extrapolation error, which may lead to unstable training and degrade agent performance \cite{ref6}. \tx{Offline RL can be divided into two parts: model-free RL and model-based RL, which take different approaches to solve the above problems.}

\tx{Model-free offline RL, independent of environment model, addresses the distribution shift challenge through two principal techniques: policy constraints \cite{ref7,ref8,ref9} and value penalization \cite{ref10,ref11}.}
The goal of the above methods is to minimize the discrepancy between the learned policy and that in the offline dataset. While they achieve remarkable performance, these methods fail to consider the effects of OOD data, hampering the exploration and performance improvement. \tx{Lyu \emph{et al}. \cite{ref12} proved that the performance can be effectively improved through exploring the OOD region. Therefore, some data augmentation techniques are studied to improve performance.}

\tx{Model-based offline RL, which trains an environment model through the offline dataset, enables the agent to continually interacting with the trained environment model to broaden data coverage and enhance exploration in the OOD region \cite{ref13}. The expansion of OOD region data can effectively mitigate distribution shift.} 
However, due to the gap between the estimated and true environment model, the generated data may be inaccurate, that is, the generated model data is inaccurate, which brings risks for agent training. The current methods solve this problem by performing conservative policy optimization. \tx{There are two types of methods to achieve conservatism. The first type involves imposing reward penalties on OOD data, either directly through uncertainty quantification \cite{ref15,ref16, ref50} or indirectly via value function regularization \cite{ref17}. The second type adopts a robust adversarial training framework, which enforces conservatism by training an adversarial environment model \cite{ref20,ref34}.}

\tx{The uncertainty quantification of the first type \cite{ref15,ref50}, estimated by neural networks, suffers from low reliability and reduces agent performance. Therefore, Yu \emph{et al}. \cite{ref17} proposed the COMBO algorithm based on CQL \cite{ref10}, that is, the value regularization of the first type, which regularizes model data to suppress unreliable data and enhance reliable offline data without estimating uncertainty. The second type of methods \cite{ref20,ref34} require continuous updates to the environment model during policy learning, leading to increased computational demands. Table \ref{tab_ref} summarizes the recent model-based offline RL methods by the type of algorithm and computation cost. In this paper, we focus on the value regularization method of the first type.}


\begin{figure*}
	\centering 
	\includegraphics[width=0.9\textwidth]{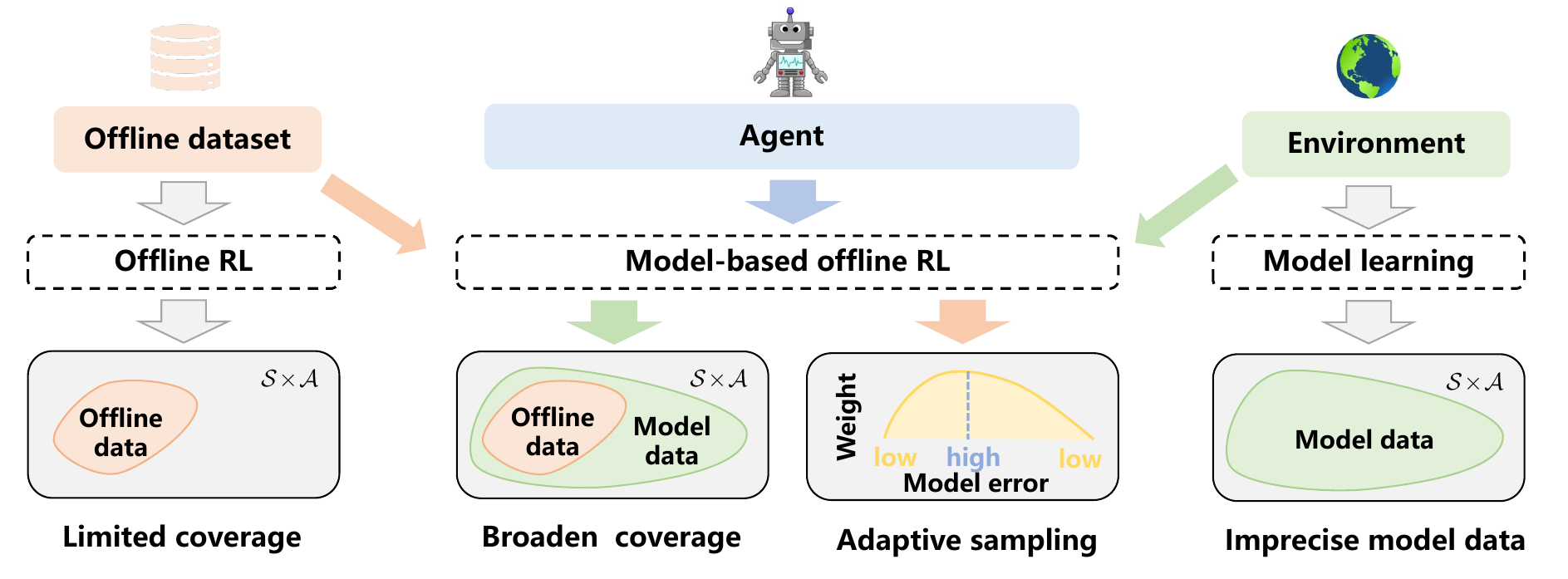}
	\caption{The framework of mildly conservative model-based offline RL. This method integrates offline RL with model-based RL, aiming to leverage the model to broaden data coverage. The key lies in designing a model data adaptive sampling distribution based on the magnitude of model errors, which enables the adaptive adjustment of model data penalty.}
	\label{fig1}
\end{figure*}

\begin{table}
	\normalsize
	\caption{The summary of recent model-based offline RL research.}
	\label{tab_ref}
	\centering
	\resizebox{9cm}{!}{
		\begin{tabular}{c|ccc|cc}
\toprule[1pt] 
\multirow{2}{*}{\textbf{Methods}}& \textbf{Uncertainty} & \textbf{Regularization} & \textbf{Adversarial}  & \textbf{Model data} & \textbf{Computation}\\ 
& \textbf{qualification}& \textbf{indirectly} & \textbf{training}& \textbf{difference} & \textbf{complexity}\\
\midrule
MOPO \cite{ref15}& \Checkmark & & &\XSolidBrush & low\\
MOReL\cite{ref16}& \Checkmark & & &\XSolidBrush & low\\
MAPLE\cite{ref50}& \Checkmark & & &\XSolidBrush & low\\
RAMBO\cite{ref20} & & &\Checkmark& \XSolidBrush& high\\
ARMOR\cite{ref34}&  & &\Checkmark& \XSolidBrush& high\\
COMBO\cite{ref17}& &\Checkmark& & \XSolidBrush& low\\
\midrule
DOMAIN& &\Checkmark& & \Checkmark& low\\
			\bottomrule[1pt]
		\end{tabular}
	}
\end{table}

\tx{For model-based offline RL, when regularizing model data to perform conservative policy, the current methods fail to consider the differences between the model data, that is, the errors ofthe  model data are different, which brings overly conservative policy.} For example, for precise model data, penalizing them as if they were imprecise model data can lead to underutilization of the precise data, thereby increasing the conservativeness of the algorithm and reducing agent performance. \tx{Motivated by this problem, we design an adaptive sampling distribution of model data, where the regularization for model data can be adjusted adaptively.} Then, a mildly conservative model-based offline RL algorithm (DOMAIN) is proposed without estimating model uncertainty (as illustrated in Fig. \ref{fig1}). Our main contributions are summarized as follows: 

\begin{enumerate}
	\item A new model-based offline RL method, named DOMAIN, is proposed, where the adaptive sampling distribution of model samples is designed to adaptively adjusts the model data penalty. 
	\item Theoretical analysis demonstrates that the value function learned by the DOMAIN in the OOD region is a lower bound of the true value function, and the DOMAIN has a safety policy improvement guarantee.
 
	\item The results of an extensive experiment indicate that DOMAIN can outperform previous RL algorithms on the D4RL benchmark.
\end{enumerate}

\tx{Here, the mild conservatism of DOMAIN is reflected as: higher exploration in the OOD region and better performance of the agent. This algorithm imposes heavier penalties on model data with high errors while reducing the penalty for accurate model data.} Notably, the exploration risk for the unknown region can be reduced through underestimating the value of inaccurate model data. Through reducing the penalty on the $Q$ value of model data with small error, the agent can choose action $a$ with higher reward $r$ by $\arg\max_aQ(s,a)$, thus improving the performance of the learned policy. 

The framework of this paper is as follows: \tx{Section \ref{sec9} provides the detailed literature for model-free and model-based offline RL.} Section \ref{sect2} introduces the fundamental theory of RL. Section \ref{sect3} shows the framework and implementation details of the DOMAIN algorithm. Section \ref{sect4} gives some relevant theoretical analysis for DOMAIN. Section \ref{sect5} presents the algorithm performance of DOMAIN on D4RL tasks. Finally, Section \ref{sect6} summarizes the entire work.

%% file: tex/09-Related.tex
\section{\tx{Related works}} \label{sec9}
\tx{This section presents a brief overview for the model-free offline RL and model-based offline RL algorithms, and the differences and advantages between our work and the previous related works.}

\tx{\textbf{Model-free Offline RL.} The current model-free offline RL algorithms take different constraints or regularization techniques to mitigate the extrapolation error. Huang \emph{et al}. \cite{ref51} proposed a mild policy evaluation method by constraining the difference between the $Q$ values of actions supported by the target policy and those of actions contained within the offline dataset. Yang \emph{et al}. \cite{ref52} treated different samples with different policy constraint intensities, which effectively improves the agent's performance. Huang \emph{et al}. \cite{r_1} proposed the de-pessimism operator to estimate $Q$ values through judging whether the actions are in the OOD region. Cao \emph{et al}. \cite{r_3} simultaneously performed behavior regularization on policy evaluation and policy improvement to improve performance. However, the main challenge in offline RL arises from the limited data coverage. Model-free offline RL algorithms rely solely on offline datasets to learn policies, which limits the agent’s ability to explore. Therefore, some model-free offline RL methods are widely studied to enhance the agent's performance.}  


\tx{\textbf{Model-based Offline RL.} Model-based offline RL generates more state-action pair data through the trained environment model and improves the exploration for the OOD region. Since the gap between the trained and the actual environment model, the current model-based offline RL algorithms adopt a penalty, regularization or adversarial training method to perform a conservative policy. Yu \emph{et al}. \cite{ref15}, Chen \emph{et al}. \cite{ref50} and Zhu \emph{et al}. \cite{r_5} estimated model uncertainty in different ways, and used model uncertainty as a reward penalty to punish model data. Yu \emph{et al}. \cite{ref17} considered the estimation errors for model uncertainty and utilized value function regularization to push down unreliable model data. Different from the above value penalty methods, Rigter \emph{et al}. \cite{ref20} and Bhardwaj \emph{et al}. \cite{ref34} used an adversarial training framework, forcing the agent to act conservatively in the OOD region. Shi \emph{et al}. \cite{r_6} and Liu \emph{et al}. \cite{r_7} incorporated a robust Markov Decision Process into policy learning to achieve conservative value estimation for model data.}

\tx{\textbf{Our work.} The proposed method, DOMAIN, belongs to model-based offline RL. The current model-based offline RL methods seldom consider the difference between the model data, that is, the precise model data is rarely fully utilized, leading to overly conservative policies. Compared with previous works \cite{ref15,ref16,ref17,ref50,ref20,ref34}, the DOMAIN adopts the adaptive sampling distribution of model data, where the penalty for high-error model data is increased and the penalty for accurate model data is reduced or even rewarded. Moreover, although training the discriminator to calculate the adaptive sampling distribution requires additional computational cost, the computational complexity of DOMAIN is lower than that of the adversarial training method \cite{ref20,ref34}. Theoretical analysis shows that DOMAIN has a policy improvement guarantee, and $Q$-values learned by DOMAIN for the OOD region are lower bounds of the true $Q$-values.} 

%% file: tex/02-Preliminaries.tex
\section{Preliminaries}\label{sect2}
\tx{In this section, we provide the basic notations for reinforcement learning algorithms, including offline RL and model-based offline RL.}

\textbf{Reinforcement Learning.}
The framework of RL is based on the Markov Decision Process (MDP) $\mathcal{M}=\left(\mathcal{S}, \mathcal{A}, r, T_{\mathcal{M}}, \rho_{0}, \gamma\right)$, where $\mathcal{S}$ is the set of states, $\mathcal{A}$ is the set of actions, $r(s, a)$ is the reward function, $T_{\mathcal{M}}\left(s^{\prime} \mid s, a\right)$ represents the dynamic system under $\mathcal{M}$, $\rho_{0}$ is the initial state distribution, and $\gamma \in(0,1)$ is the discount factor \cite{ref5}. The goal of RL is to learn a policy that maximizes the cumulative reward $J(\mathcal{M}, \pi)=\mathbb{E}_{(s, a) \sim d_{\mathcal{M}}^{\pi}(s, a)}[r(s, a)] / 1-\gamma$, where $d_{\mathcal{M}}^{\pi}(s, a)=d_{\mathcal{M}}^{\pi}(s) \pi(a \mid s)$ is the marginal distribution of states and actions under the learned policy $\pi$, and $d_{\mathcal{M}}^{\pi}(s)$ is the discounted marginal state distribution under the learned policy $\pi$.

Considering the Bellman operator $\mathcal{T}^{\pi} Q(s, a):=r(s, a)+\gamma \mathbb{E}_{s^{\prime} \sim T_{\mathcal{M}}\left(s^{\prime} \mid s, a\right)}\left[\max_{a'\in \mathcal{A}}Q^{\pi}\left(s^{\prime}, a^{\prime}\right)\right]$, its sample-based counterpart is $\widehat{\mathcal{T}}^{\pi} Q(s, a):=r(s, a)+\gamma \max_{a'\in\mathcal{A}}Q\left(s^{\prime}, a^{\prime}\right)$\cite{ref17}. The Actor-Critic algorithm minimizes the Bellman error to learn the $Q$-function and maximizes the $Q$-function to optimize the policy $\pi$. The formulas are expressed as follows:
\begin{equation}\label{eq1}
	\begin{aligned}
		&\widehat{Q} \leftarrow \arg \min _{Q} \mathbb{E}_{s, a, s^{\prime} \sim \mathcal{D}}\Big[\left(Q(s, a)-\widehat{\mathcal{T}}^{\pi} \widehat{Q}(s, a)\right)^{2}\Big]\\
		&\widehat{\pi} \leftarrow \arg \max _{\pi} \mathbb{E}_{s \sim \mathcal{D}, a \sim \pi}\left[\widehat{Q}(s, a)\right]
	\end{aligned}
\end{equation}
where $\mathcal{D}$ can be a fixed offline dataset or replay buffer generated by the current policy $\widehat{\pi}$ interacting with the environment.

\textbf{Offline RL.}
The goal of offline RL is to learn a policy $\pi_{\theta}(\cdot \mid s)$ from an offline dataset $\mathcal{D}=\left\{\left(s_{i}, a_{i}, r_{i}, s_{i}^{\prime}\right)\right\}_{i=1}^{n}$ without interacting with the environment, to maximize the expected discounted reward. However, there exists the distribution shift between the learned policy $\pi_{\theta}(\cdot \mid s)$ and the behavioral policy $\pi_{b}(\cdot \mid s)$ of the dataset, leading to unstable training of the agent \cite{ref5}. To alleviate this issue, previous studies \cite{ref7,ref8,ref9,ref10,ref11} have introduced regularization on the value function of OOD actions to reduce the probability of selecting OOD actions. Here primarily focuses on CQL\cite{ref10}. The formulas for policy improvement and policy evaluation are as follows:
\begin{equation}\label{eq2}
	\begin{aligned}
		\widehat{Q} \leftarrow & \arg \min _{Q} \frac{1}{2} \mathbb{E}_{s, a, s^{\prime} \sim \mathcal{D}}\Big[\left(Q(s, a)-\widehat{\mathcal{T}}^\pi \widehat{Q}(s, a)\right)^2\Big]\\
		&+\beta\Big \{\mathbb{E}_{s \sim \mathcal{D}, a \sim \mu(\cdot \mid s)}\left[Q(s, a)\right]-\mathbb{E}_{s, a \sim \mathcal{D}}\left[Q(s, a)\right]\Big \}\\
		\widehat{\pi} \leftarrow &\arg \max _{\pi} \mathbb{E}_{s \sim \mathcal{D}, a \sim \pi_{k}}\left[\widehat{Q}(s, a)\right]
	\end{aligned}
\end{equation}
where $\beta$ represents the regularization coefficient, $\mu(\cdot \mid s)$ denotes the distribution from which actions are sampled, which can be a uniform distribution or a learned policy. CQL effectively penalizes the $Q$-values in the OOD region, providing a conservative estimate of the value function and alleviating the challenges of overestimation and distribution shift.

\tx{\textbf{Model-based Offline RL.} Model-based offline RL involves two steps: environment model learning and policy learning. It uses the learned environment model $\hat{T}$ to generate more model data, which can broaden data coverage and can further enhance the exploration for the OOD region. The environment model $\widehat{T}_{\phi}\left(s^{\prime}, r \mid s, a\right)$ is trained using the offline dataset, and then, the policy is learned based on both offline data and generated model data. Following the previous work \cite{ref15,ref16}, the environment model is parameterized as Gaussian distribution $\widehat{T}_{\phi}(s^{\prime}, r \mid s, a)=\mathcal{N}[\mu_{\phi}(s, a), \Sigma_{\phi}(s, a)]$, which can better describe the randomness of the real environment.} The loss function for training the environment model employs maximum likelihood estimation, which can be denoted as: $\mathcal{L}_{\mathcal{M}}(\phi)=-\mathbb{E}_{\left(s, a, r, s^{\prime}\right) \sim \mathcal{D}}[\log\widehat{T}_{\phi}(s^{\prime},r| s, a)]$. With this learned model, an estimated MDP $\widehat{\mathcal{M}}$ can be obtained. 

Subsequently, the agent interacts with the learned policy $\widehat{\pi}$ and the environment model $\widehat{T}_{\phi}$ to generate model data $\mathcal{D}_{\text {model }}$ through $H$-step rollouts. \tx{The policy learning is then performed using the combined dataset $\mathcal{D}_{\text {offline }} \cup \mathcal{D}_{\text {model }}$, where the data is sampled from model data $\mathcal{D}_{\text {model }}$ with the probability $f, f \in[0,1]$, and from the offline data $\mathcal{D}_{\text {offline }}$ with the probability $1-f$.} However, offline dataset fails to cover the entire state-action space, resulting in disparities between the learned environment model $\widehat{T}_{\phi}\left(s^{\prime}, r \mid s, a\right)$ and the actual environment model $T\left(s^{\prime}, r \mid s, a\right)$. Similar to \cite{ref19,ref20}, we employed a model ensemble approach to reduce biases introduced by probabilistic networks.

%% file: tex/03-Method.tex
\section{ Methodology}\label{sect3}


\tx{This section provides a detailed description of the mildly conservative model-based offline reinforcement learning (DOMAIN) algorithm, including the mildly conservative estimation of the value function and adaptive value regularization on model data.}

\subsection{Mildly Conservative Value Estimation}
Conservative policy evaluation within the DOMAIN algorithm framework can be described as follows: Given an offline dataset $\mathcal{D}_{\text {offline }}=\left\{\left(s_{i}, a_{i}, r_{i}, s_{i}^{\prime}\right)\right\}_{i=1}^{n}$, a learned policy $\widehat{\pi}$, and a learned environmental model $\widehat{T}_{\phi}$, the primary objective is to obtain a conservative estimate of $Q(s, a)$. \tx{To achieve this, the policy evaluation step in DOMAIN utilizes the CQL framework (see Eq. \eqref{eq2})}. This principle can be mathematically expressed as
\begin{equation}\label{eq3}
	\begin{aligned}
		\min _{Q} \max _{\rho} \mathcal{L}_{\beta}\left(Q, \widehat{\mathcal{T}}^{\pi} \widehat{Q}\right)+\lambda\Big\{\mathbb{E}_{s, a \sim \rho(s, a)}\big[Q(s, a)\big]\\
		-\mathbb{E}_{s, a \sim \mathcal{D}_{\text {offline }}}\big[Q(s, a)\big]+\mathcal{H}(\rho)\Big \}
	\end{aligned}
\end{equation}
where $\lambda$ is a scaling factor used to adjust the weight of the regularization term, $\rho(s, a)$ denotes the sampling distribution of the model data, and $\mathcal{H}(\rho)$ is a constraint term for $\rho(s, a)$. The term $\mathbb{E}_{s, a \sim \rho(s, a)}[Q(s, a)]$ is employed to penalize unreliable model data, while $\mathbb{E}_{s, a \sim \mathcal{D}_{\text {offline }}}[Q(s, a)]$ is used to reward reliable offline data. The term $\mathcal{L}_{\beta}(Q, \widehat{\mathcal{T}}^{\pi} \widehat{Q})$ represents the Bellman error. 

In the Actor-Critic framework, the goal of policy evaluation is to minimize the Bellman error \cite{ref35}. In the DOMAIN algorithm, the Bellman error, consisting of the Bellman error from the offline data and the model data, can be denoted as
\begin{equation}\label{eq5}
	\begin{aligned}
		\mathcal{L}_{\beta}&\left(Q, \widehat{\mathcal{T}}^{\pi} \widehat{Q}\right)\\
		=\Big \{&(1-f) \mathbb{E}_{s, a, s^{\prime} \sim \mathcal{D}_{\text {offline }}}\left[\left(Q-\mathcal{T}_{\overline{\mathcal{M}}}^{\pi} \widehat{Q}\right)(s, a)\right]^{2} \\
		&+f \mathbb{E}_{s, a, s^{\prime} \sim \mathcal{D}_{\text {model }}}\left[\left(Q-\mathcal{T}_{\widehat{\mathcal{M}}}^{\pi} \widehat{Q}\right)(s, a)\right]^{2}\Big \}
	\end{aligned}
\end{equation}
\tx{where $f$ can be determined by two guidelines. One is the error of model data, and the other is the algorithm itself. If the model data is accurate and the algorithm can effectively penalize model data with large errors, $f$ can be set larger. On the contrary, if the error of model data is large and the algorithm fails to effectively penalize them, $f$ needs to be set smaller to guarantee the stability and security of the agent.} $\widehat{\mathcal{T}}^{\pi}$ represents the actual Bellman operator used in the algorithm, while $\mathcal{T}_{\overline{\mathcal{M}}}^{\pi}$ and $\mathcal{T}_{\widehat{\mathcal{M}}}^{\pi}$ represent the operator applied under the empirical MDP and the learned model, respectively.

The value function $Q$ and the policy $\pi$ in Eq. \eqref{eq5} can be approximated using neural networks, denoted as $Q_{\omega}$ and $\pi_{\theta}$, respectively. \tx{The value function update follows the SAC framework \cite{ref21}, and consequently, the Bellman operator can be further expressed as:}
\begin{equation}\label{eq6}
	\begin{aligned}
		\widehat{\mathcal{T}}^{\pi} \widehat{Q}(s, a) \approx r+\gamma\Big(\min _{i=1,2} Q_{\omega_{i}}(s^{\prime}, a^{\prime})-\alpha \log \pi_{\theta}\Big)
	\end{aligned}
\end{equation}
\tx{where $\alpha$ is the entropy regularization coefficient, which determines the relative importance of entropy. By employing a mildly conservative policy evaluation, the policy improvement scheme can be derived:}
\begin{equation}\label{eq7}
	\begin{aligned}
		\pi_{\theta}=\arg \max _{\theta} \mathbb{E}_{s \sim \mathcal{D}, a \sim \pi_{\theta}}\Big[\min _{i=1,2} Q_{\omega_{i}}(s, a)-\alpha \log \pi_{\theta}\Big]
	\end{aligned}
\end{equation}

Here, entropy regularization is employed to prevent policy degradation. The two $Q$ value networks are utilized to select the minimum value to address overestimation in agent updates\cite{ref33}. In Eq. \eqref{eq3}, the constraint term $\mathcal{H}(\rho)$ for $\rho(s, a)$ is typically approximated using KL divergence to approach the desired sampling distribution $\omega(s, a)$, i.e., $\mathcal{H}(\rho)=-D_{K L}(\rho \| \omega)$. Thus, the following optimization problem can be formulated:
\begin{equation}\label{eq4}
	\begin{aligned}
		&\max _{\rho} \mathbb{E}_{s, a \sim \rho(s, a)}[Q(s, a)]-D_{K L}(\rho \| \omega) \\
		&\text { s.t. } {\sum_{s,a}\rho(s, a) =1, \rho(s, a) \geq 0}
	\end{aligned}
\end{equation}

The above optimization problem can be solved using the Lagrange method \cite{ref10}, yielding the following solution:
\begin{equation}\label{eq8}
	\rho^{*}(s, a)=\omega(s, a) \exp [Q(s, a)] / Z
\end{equation}
where $Z$ is the normalization factor. \tx{Substituting the solution obtained by solving Eq. \eqref{eq4} back into Eq. \eqref{eq3}, the following expression can be derived:}
\begin{equation}\label{eq9}
	\begin{aligned}
		\min _{Q} \mathcal{L}_{\beta}\left(Q, \widehat{\mathcal{T}}^{\pi} \widehat{Q}\right)+\lambda\Big \{\log \mathbb{E}_{s, a \sim \omega(s, a)} \exp \left[Q(s, a)\right]\\
		-\mathbb{E}_{s, a \sim \mathcal{D}}\left[Q(s, a)\right]\Big \}
	\end{aligned}
\end{equation}

By optimizing Eq. \eqref{eq9}, a mildly conservative value estimation can be achieved. The sampling distribution of model data $\omega(s, a)$ reflects the degree of penalization for model data. In below part, we give further derivation for the sampling distribution of model data $\omega(s, a)$.

\subsection{Adaptive Value Regularization on Model Data}
\tx{When performing policy evaluation, we aim to assign higher penalty coefficients to larger model data errors, thereby increasing the value for $\omega(s, a)$. Drawing inspiration from \cite{ref22,bbbbb_3}, the KL divergence is employed to quantify the magnitude of model data errors:}
\begin{equation}\label{eq10}
	\begin{aligned}
		g(s, a)=\mathbb{E}_{s^{\prime} \sim T_{\widehat{\mathcal{M}}}\left(s^{\prime} \mid s, a\right)}\left[\frac{T_{\widehat{\mathcal{M}}}\left(s^{\prime} | s, a\right)}{T_{\mathcal{M}}\left(s^{\prime} | s, a\right)}\right]
	\end{aligned}
\end{equation}
\tx{where $T_{\widehat{\mathcal{M}}}$ represents the state transition model in the learned environment model $\widehat{\mathcal{M}}$, while $T_{\mathcal{M}}$ corresponds to that in the true environment model $\mathcal{M}$.} Since sampling distribution for model data needs to satisfy the equation: $\sum_{s,a}\omega(s, a) =1$, the sampling distribution $\omega(s, a)$ based on finite model dataset can be denoted as:
\begin{equation}\label{eq11}
	\omega(s, a)={g(s, a)}\big/{\sum_{s',a'} g\left(s^{\prime}, a^{\prime}\right)}
\end{equation}

Hence, to design an accurate sampling distribution, it is crucial to compute the dynamic ratio value $T_{\widehat{\mathcal{M}}} / T_{\mathcal{M}}$ precisely. Following \cite{ref22}, the following equation can be derived by applying Bayes' theorem:
\begin{equation}\label{eq12}
	\begin{aligned}
		\frac{T_{\widehat{\mathcal{M}}}\left(s^{\prime}|s, a\right)}{T_{\mathcal{M}}\left(s^{\prime}|s, a\right)}=\frac{p\left(\text { model }|s, a, s^{\prime}\right) p\left(\text { offline }|s, a\right)}{p\left(\text { offline }|s, a, s^{\prime}\right) p\left(\operatorname{model}|s, a\right)}
	\end{aligned}
\end{equation}
where $p\left(\text{model}|s, a, s^{\prime}\right)$ denotes the probability that the pair $\left(s, a, s'\right)$ belongs to model data. To estimate the above probabilities, two discriminators are constructed, namely $D_{\Phi_{s a s}}\left(\cdot \mid s, a, s^{\prime}\right)$ and $D_{\Phi_{s a}}\left(\cdot \mid s, a\right)$, which approximate the probabilities using parameters $\Phi_{s a s}$ and $\Phi_{s a}$, respectively. The network loss function employed for training the discriminators is the cross-entropy loss \cite{ref23}:
\begin{equation}\label{eq13}
	\begin{aligned}
		\mathcal{L}(\Phi_{sas})=&-\mathbb{E}_{\mathcal{D}_{\text {offine }}}\big[\log D_{\Phi_{sas }} \text { (offline }|s, a, s^{\prime})\big]\\
		&-\mathbb{E}_{\mathcal{D}_{\text {model }}}\big[\log D_{\Phi_{sas}}(\text { model }|s, a, s^{\prime})\big] \\
		\mathcal{L}(\Phi_{s a})=&-\mathbb{E}_{\mathcal{D}_{\text {offline }}}\big[\log D_{\Phi_{s a}}(\text { offline }|s, a)\big]\\
		&-\mathbb{E}_{\mathcal{D}_{\text {model }}}\big[\log D_{\Phi_{s a}}(\text { model } | s, a)\big]
	\end{aligned}
\end{equation}

\tx{By training the discriminator networks, the gap between the learned and actual environment can be estimated. Subsequently, the sampling distribution of model data can be obtained. Based on this, the varying sampling probabilities among the model data can introduce adaptive value regularization for the model data.}

\subsection{Algorithm Details of DOMAIN}
DOMAIN imposes heavier penalty on model data with high errors while reducing the penalty for accurate model data to alleviate conservatism. 
By reducing the penalty on the $Q$ value of model data with small errors, the agent may choose action $a$ with higher reward $r$ through $\arg\max_aQ(s,a)$, thus improving the performance of the learned policy. The algorithm DOMAIN, summarized in Algorithm \ref{alg:algorithm1}, is based on SAC \cite{ref21} and MOPO \cite{ref15}. $N$ dynamic environment models are trained and a subset of $M$ models with superior training accuracy is selected. Eq.s \eqref{eq6} and \eqref{eq7} adhere to SAC framework, with the parameter $\alpha$ being automatically adjusted \cite{ref21}. 

DOMAIN employs Eq. \eqref{eq9} to perform mildly conservative policy evaluation, Eq. \eqref{eq7} for policy improvement, and Eq. \eqref{eq13} to train the discriminator. \tx{The term $g(s, a)$ represents the expectation of the dynamic ratio, which is estimated by sampling $m$ states from a Gaussian distribution $\widehat{T}_{\phi}\left(s^{\prime}, r \mid s, a\right)=\mathcal{N}[\mu_{\phi}(s, a), \sum_{\phi}(s, a)]$ obtained from the trained model. The weight $\lambda$ of the regularization term and the rolling horizon $H$ have a substantial influence on the agent training.} The detailed analysis can be found in Section \ref{Parameter Analysis}.

\begin{figure}[!t]
	\centering
	\renewcommand{\algorithmicrequire}{\textbf{Input:}}
	\renewcommand{\algorithmicensure}{\textbf{Output:}}
	\removelatexerror
	\begin{algorithm}[H]
		\label{alg:algorithm1}
		\caption{DOMAIN: Mildly Conservative Model-Based Offline Reinforcement Learning}
		\begin{algorithmic}[1]
			\REQUIRE offline dataset $\mathcal{D}_{\text{offline}}$, critic $Q_{\omega_1}$, $ Q_{\omega_2}$,  policy $\pi_\theta$, discriminator $ D_{\Phi_{s a s}}, D_{\Phi_{s a}}$, environment model $\widehat{T}_\phi$, rollout policy $ \mu(\cdot \mid s) $ and rollout length $ H $. 
			\ENSURE learned policy $\pi_\theta(\cdot \mid s)$.
			
			\STATE \textbf{Initialization}: target network $ Q_{\overline{\omega}_{1,2}} \leftarrow Q_{\omega_{1,2}}$,  and model dataset $\mathcal{D}_{\text{model }}=\varnothing$.
			
			\STATE Use $\mathcal{D}_{\text{offline} }$  to train $N $ environment models $\{\widehat{T}_\phi^i\}_{i=1}^N$.
			
			\FOR {$t=1,2,3 \cdots, T $}
			\STATE Collect model date: rollout $(\mu(\cdot \mid s), \widehat{T}_\phi, H)$, and add rollout date (model data) into $\mathcal{D}_{\text{model}}$ ;
			\STATE Update discriminator $D_{\Phi_{s a s}}, D_{\Phi_{s a}}$  by Eq. \eqref{eq13}, and calculate adaptive sampling distribution $\omega$;
			\STATE Update policy $\pi_\theta$ and critic $Q_{\omega_{1,2}}$ through Eq. \eqref{eq7} and Eq. \eqref{eq9} (See Table \ref{tab_1} for related parameters);
			\STATE \textbf{if} $t$ \% update period = 0:
			\STATE \quad Soft update: $\overline{\omega}_{1,2} \leftarrow \tau \omega_{1,2}+(1-\tau) \overline{\omega}_{1,2} $;
			\STATE \textbf{end if}
			\ENDFOR
		\end{algorithmic}
	\end{algorithm}
\end{figure}

\section{Theoretical Analysis of DOMAIN}\label{sect4}
In this section, we demonstrate that the adaptive sampling distribution of model data is equivalent to the adaptive adjustment of penalties for rewards. \tx{Furthermore, we prove that the learned value function in DOMAIN serves as the lower bound for the true value in the OOD region, effectively suppressing the value under the highly uncertain region. By employing a lower-bound optimization policy, DOMAIN has a safety policy improvement guarantee.}

\subsection{Adaptive Adjustment of Penalties for Rewards}
Before the theoretical analysis, we give the following theorem, which proves that the Bellman operator $\widehat{\mathcal{T}}^\pi$ can converge to a specific $Q$ value.
\newtheorem{theorem}{\bf Theorem}
\begin{theorem}\label{thm4}
Let $\|\cdot\|_\infty$ is the $\mathcal{L}_\infty$ norm and ($\mathbb{R}^{|\mathcal{S} \times \mathcal{A}|}, \|\cdot\|_\infty$) is complete space. Then, the Bellman operator $\widehat{\mathcal{T}}^\pi$ is a $\gamma$-contraction mapping operator, i.e. $\|\widehat{\mathcal{T}}^\pi Q_1-\widehat{\mathcal{T}}^\pi Q_2\|_{\infty}\leq \gamma\left\| Q_1- Q_2\right\|_{\infty}$.
\end{theorem}

The detailed proof can be found in Appendix A (Please see Supplementary Material). Theorem \ref{thm4} shows that the operator $\widehat{\mathcal{T}}$ is a $\gamma$-contraction mapping in the $\mathcal{L}_\infty$ norm. According to the fixed point theorem, any initial $Q$ function can converge to a unique fixed point $Q^*$ by repeatedly applying the Bellman operator $\widehat{\mathcal{T}}^\pi$. 

Then, by setting the derivative of the Eq. \eqref{eq9} concerning $Q^{k}$ to zero during the $k$-th iteration, the following equation can be obtained:
\begin{equation}\label{eq14}
	\widehat{Q}^{k+1}(s, a)=\left(\widehat{\mathcal{T}}^{\pi} \widehat{Q}^{k}\right)(s, a)-\lambda\left[\frac{\omega(s, a)-d(s, a)}{d_{\beta}(s, a)}\right]
\end{equation}
where the distribution for offline data sampling, denoted as $d(s, a)$, is defined as $d(s, a)=d^{\pi_{b}}(s, a)=d^{\pi_{b}}(s) \pi_{b}(a|s)$, and $d^{\pi_{b}}(s, a)$ represents the marginal distribution of states and actions under the behavioral policy $\pi_{b}$. The data sampling distribution used for computing the Bellman error, denoted as $d_{\beta}(s, a)$, is defined as $d_{\beta}(s, a)=(1-f) d^{\pi_{b}}(s, a)+f d_{\widehat{\mathcal{M}}}^{\pi}(s, a)$. Here, $d_{\widehat{\mathcal{M}}}^{\pi}(s, a):=d_{\widehat{\mathcal{M}}}^{\pi}(s) \pi(a|s)$ represents the marginal distribution of states and actions under the learning policy $\pi$, and $d_{\widehat{\mathcal{M}}}^{\pi}(s)$ represents the discounted marginal state distribution when executing policy $\pi$ in the learned model $\widehat{\mathcal{M}}$.

We denote the second term in Eq. \eqref{eq14} as $\eta(s, a)$, which serves as the adjustment of penalties for rewards. \tx{When the model data $(s, a)$ exhibits significant errors, the value of the designed sampling distribution $\omega(s, a)$ increases. This suggests that the data point $(s, a)$ either lies in the OOD region or a region with a relatively low density within the offline dataset, implying that $d(s, a)$ approaches zero.} Consequently, $\omega(s, a)>d(s, a)$, resulting in $\eta(s, a)>0$ that serves as a penalty term. Moreover, the magnitude of penalization in the \emph{Q} value increases with larger errors in the model data. In contrast, when the model data error is small, that is the data is situated in a region proximal to true environment, it is possible for $\omega(s, a)$ to be smaller than $d(s, a)$, leading to $\eta(s, a)<0$ which acts as a reward term. This encourages exploration in the accurate model data region during the training process. 

\subsection{DOMAIN Optimizes a Low Bound Value}
\tx{Considering the influence of sampling and model errors, we present the conditions for the existence of a lower bound on the value function in DOMAIN and demonstrate that DOMAIN does not underestimate the value function for any state.} Additionally, we discuss a sufficient condition under which the lower bound of DOMAIN is tighter than that of COMBO, indicating that DOMAIN exhibits weaker conservatism.

\begin{theorem}\label{thm1}
	The value function learned using Eq. \eqref{eq9} serves as a lower bound on the true value function, i.e. $V^{\pi}(s) \geq \widehat{V}^{\pi}(s)$, given that the following conditions are satisfied: $\sum_a \omega(s, a) >\xi(s) \sum_a d(s, a) $ and $\lambda \geq \delta_{l}$, where the state coefficient $\xi(s)$ and $\delta_{l}$ are defined as:
 \begin{equation*}
     \xi(s)=\frac{(1-f) d^{\pi_{b}}(s) \max\limits _{a}\left[\pi_{b}(a \mid s) / \pi(a \mid s)\right]+f d_{\widehat{\mathcal{M}}}^{\pi}(s)}{(1-f) d^{\pi_{b}}(s) \min \limits_{a}\left[\pi_{b}(a \mid s) / \pi(a \mid s)\right]+f d_{\widehat{\mathcal{M}}}^{\pi}(s)} \geq 1 
 \end{equation*}
	\begin{equation}\label{eq15}
		\begin{aligned}
			&\delta_{l}=\frac{(1-f) \frac{C_{r, T, \delta} R_{\max }}{(1-\gamma) \min _{s, a}\{\sqrt{|\mathcal{D}|}\}}}{\min _{s}\left\{\frac{(\xi(s)-1) \sum_{a} d(s, a)  }{\left[(1-f) d^{\pi_{b}}(s) \max _{a}\left\{\frac{\pi_{b}(a \mid s)}{\pi(a \mid s)}\right\}+f d_{\widehat{\mathcal{M}}}^{\pi}(s)\right]}\right\}}\\
			&+\frac{f[\max \limits_{s, a}\left\{\left|r_{\mathcal{M}}-r_{\mathcal{M}}\right|\right\}+\frac{2 \gamma R_{\max }}{1-\gamma} \max\limits _{s, a}\left\{D_{T V}\left(T_{\mathcal{M}}, T_{\widehat{\mathcal{M}}}\right)\right\}]}
			{\min _{s}\left\{\frac{(\xi(s)-1) \sum_a d(s, a) } {\left[(1-f) d^{\pi_{b}}(s) \max _{a}\left\{\frac{\pi_{b}(a \mid s)}{\pi(a \mid s)}\right\}+f d_{\widehat{\mathcal{M}}}^{\pi}(s)\right]}\right\}}
		\end{aligned}
	\end{equation}
\end{theorem}

\tx{The detailed proof is provided in Appendix B. Based on the theorem, for a sufficiently large $\lambda$ and when the aforementioned conditions are met, $V^{\pi}(s) \geq \widehat{V}^{\pi}(s)$ holds. when a sufficient amount of data is available, $\delta_{l}$ primarily is primarily attributed to model errors, $C_{r, T, \delta} R_{\max } /((1-\gamma) \min _{s, a}\{\sqrt{|\mathcal{D}|}\})$ approaches zero. Therefore, when the model error is small and the model data ratio $f$ is appropriately chosen, a sufficiently small $\lambda$ can ensure $V^{\pi}(s) \geq \widehat{V}^{\pi}(s)$.}

Furthermore, DOMAIN does not underestimate the value function for all states $s$. The value function is underestimated only when the condition in Eq. \eqref{eq15} is satisfied. However, there is a possibility of overestimating the data with high probabilities in the distribution $d(s, a)$ of the offline data set. Since the data points $(s, a)$ with a higher frequency of occurrence in region $\mathcal{D}_{\text {offline }}$ are considered as in-distribution data, the learned environment model $\widehat{T}_{\phi}$ exhibits smaller prediction errors for such data, resulting in smaller values of $\omega(s, a)$. As a result, it becomes challenging to satisfy the condition $\sum_a \omega(s, a) >\xi(s) \sum_a d(s, a)$, leading to overestimation. This aligns with the adaptive adjustment of penalties for reward.
\vspace{4pt}

\begin{theorem}\label{thm2}
	Let $\kappa_{1}$ and $\kappa_{2}$ denote the average value functions of the DOMAIN and COMBO methods, respectively. Specifically, $\kappa_{1}=\mathbb{E}_{s \sim d(s), a \sim \pi(a \mid s)}[\widehat{Q}_{1}^{\pi}]$, $\kappa_{2}=\mathbb{E}_{s \sim d(s), a \sim \pi(a \mid s)}[\widehat{Q}_{2}^{\pi}]$. It holds that $\kappa_{1}>\kappa_{2}$ if the following condition is satisfied:
	\begin{equation}\label{eq16}
		\mathbb{E}_{s \sim d(s)}\Big[d_{\widehat{\mathcal{M}}}^{\pi}(s)\Big] \geq \mathbb{E}_{s \sim d(s), a \sim \pi(a \mid s)}\Big[\omega(s, a)\Big]
	\end{equation}
\end{theorem}
where $\widehat{Q}_{1}^{\pi}$ and $\widehat{Q}_{2}^{\pi}$ are the $Q$-value function of DOMAIN and COMBO in learned policy $\pi$ respectively. $d(s)$ represents the state distribution of any data set, and $d_{\widehat{\mathcal{M}}}^{\pi}(s)$ represents the distribution of states under the learned model. 

The detailed proof of this theorem can be found in Appendix C. When the dataset is the offline dataset, the data points $(s, a)$ with higher densities in $d(s, a)$ exhibit smaller model prediction errors, which implies $w(s,a)$ approaches zero. The generated model data is more likely to be concentrated in-distribution region, indicating $d_{\widehat{\mathcal{M}}}^{\pi}(s)$ is large. Hence, Eq. \eqref{eq16} is readily satisfied, indicating that the conservatism of DOMAIN is weaker than that of COMBO.

\subsection{DOMAIN Guarantees the Safety of Policy Improvement}
The offline dataset is the only dataset available for policy learning of DOMAIN, and the policy in the offline dataset can be described as behavior policy $\pi_{b}(a \mid s)$. \tx{The goal of DOMAIN is to learn an optimal policy $\pi^{*}(a\mid s)$ in the actual MDP $\mathcal{M}$ through maximizing the cumulative return $J(\mathcal{M}, \pi):=\mathbb{E}_{(s, a) \sim d_{\mathcal{M}}^{\pi}(s, a)}[r(s, a)] / 1-\gamma$, and the performance gap between agent under optimal policy and behavioral policy should be bounded.} Building upon previous works \cite{ref12,ref17,ref26,ref27}, we prove that the DOMAIN provides a $\zeta$-safe policy improvement for the behavioral policy.
\vspace{6pt}

\begin{theorem}\label{thm3}
	Let $\pi^{*}(a \mid s)$ be the policy optimized by DOMAIN. Then, $\pi^{*}(a \mid s)$ represents a $\zeta$-safe policy improvement over $\pi_{b}(a \mid s)$ in the actual MDP $\mathcal{M}$, satisfying $J\left(\mathcal{M}, \pi^{*}\right)-J\left(\mathcal{M}, \pi_{b}\right) \geq \zeta$ with high probability $1-\delta$. The value of $\zeta$ is given by:
	\begin{equation}\label{17}
		\begin{aligned}
			\zeta=&\underbrace{\frac{\lambda}{1-\gamma}\Big[\varpi\left(\pi^{*}, f\right)-\varpi\left(\pi_{b}, f\right)\Big]}_{:=\mu_{1}}\\
			&-\underbrace{2 \frac{1-f}{1-\gamma} \mathbb{E}_{(s, a) \sim d_{\mathcal{M}}^{\pi^*}}\left[\frac{C_{r, \delta}+R_{\max } C_{T, \delta}}{\sqrt{|\mathcal{D}(s, a)|}}\right]}_{:=\mu_{2}} \\
			& -\underbrace{2 \frac{f}{1-\gamma} \mathbb{E}_{(s, a) \sim d_{\mathcal{M}}^{\pi^{*}}}\Big[\varepsilon_{r}(s, a)+R_{\max } D(s, a)\Big]}_{:=\mu_{3}}
		\end{aligned}
	\end{equation}
\end{theorem}
  
\begin{figure}
	\centering
	\includegraphics[width=8.6cm]{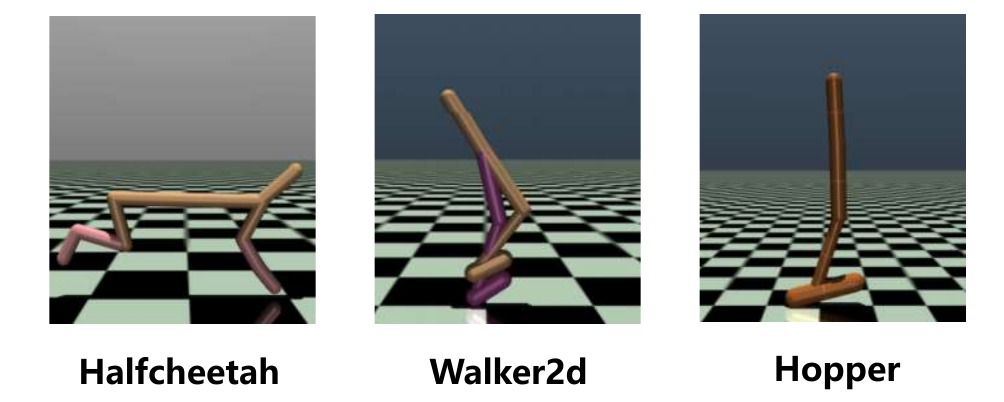}
	\caption{The description of Mujoco tasks, including Halfcheetah, Hopper, and Walker2d tasks.}
	\label{fig_2}
\end{figure}

The proof of this theorem can be found in Appendix D. Here, $\mu_{2}$ represents the sampling error term, which becomes negligible when the data volume is sufficiently large. \tx{The term $\mu_{3}$ denotes the model error term, where a larger value indicates a greater discrepancy between the trained environment model $\widehat{T}_{\phi}$ and the true model $T$. Additionally, $\varpi(\pi, f)=\mathbb{E}_{(s, a) \sim d_{\mathcal{M}_{f}(s, a)}^{\pi}}[\eta(s, a)]$ quantifies the expected rewards and penalties under a specific policy and model data ratio, capturing the difference in the expected penalty between different policies.} Therefore, $\mu_{1}$ represents the discrepancy in the expected penalties under different policies.

Since the environment model is trained using an offline dataset $\mathcal{D}_{\text {offline }}$ generated by the behavioral policy $\pi_{b}(a \mid s)$, the learned model approximates the model data distribution $d_{\widehat{\mathcal{M}}}^{\pi_{b}}$ generated by the interaction between the behavioral policy and the environment, which is close to the distribution $d^{\pi_{b}}$ of the offline dataset. Consequently, the generated model data exhibits high precision, resulting in smaller values of $\omega(s, a)$. Considering that the value of $D_{K L}\left(\pi^{*} \| \pi_{b}\right)$ is not expected to be large, the distributions $d_{\widehat{\mathcal{M}}}^{\pi_{b}}$ and $d_{\mathcal{M}}^{\pi_{b}}$ can be approximated as equal, leading to a higher expected value of $\varpi\left(\pi^{*}, f\right)-\varpi\left(\pi_{b}, f\right)>0$ with high probability. Thus, by selecting a sufficiently large $\lambda$, it can be ensured that $\zeta>0$. Furthermore, by choosing an appropriate value for $f$, even a smaller $\lambda$ can guarantee the safety improvement of the behavioral policy.

%% file: tex/04-Experiment.tex
\section{Experiments}\label{sect5}

This section aims to evaluate the performance of DOMAIN on standard benchmarks, and \tx{gives the details of experimental setup, results on D4RL tasks, computation complexity, robust performance, and key parameters analysis.}

\begin{table}
	\centering
	\caption{Base hyperparameter configuration of DOMAIN.}
	\label{tab_1}
	\renewcommand{\arraystretch}{1.2}
	\resizebox{8.5cm}{!}{
		\begin{tabular}{p{0.02cm}p{0.02cm}p{0.25cm}ll}
			\toprule[1pt] 
			&\multicolumn{3}{l}{\quad \textbf{Parameter}}  & \textbf{Value} \\
			\hline
			\multirow{4}{*}{\rotatebox{90}{\textbf{Shared}}}&\multirow{4}{*}{\rotatebox{90}{\textbf{Parameter}}}&&Number of hidden units per layer& 256/200\\
			&&& Number of iterations& 1M\\
			&&&Batch size & 256 \\
			&&&Optimizer & Adam\\
			\hline
			
			\multirow{5}{*}{\rotatebox{90}{\textbf{Environment}}}&\multirow{5}{*}{\rotatebox{90}{\textbf{Model}}}&&Model learning rate & 1e-4 \\
			&&&Number of hidden layers & 4 \\
			&&&Number of model networks $N$ & 7 \\
			&&&Number of elites $M$ & 5 \\
			&&&Ratio of model date $f$ & 0.5 \\
			\hline
			
			\multirow{5}{*}{\rotatebox{90}{\textbf{Policy}}}&\multirow{5}{*}{\rotatebox{90}{\textbf{Learning}}}&
			&Learning rate (policy/critic) & 1e-4/3e-4 \\
			&&&Number of hidden layers & 2 \\
			&&&Discount factor $\gamma$ & 0.99 \\
			&&&Soft update parameter $\tau$ & 5e-3 \\
			&&&Target entropy & -dim(A) \\
			\hline
			
			\multirow{5}{*}{\rotatebox{90}{\textbf{Discriminator}}}&\multirow{5}{*}{\rotatebox{90}{\textbf{Training}}}&&Discriminator learning rate & 3e-4 \\
			&&&Number of hidden layers & 1 \\
			&&&KL Divergence clipping range&$[10^{-45},10]$\\
			&&&Dynamics ratio clipping range & $[10^{-45},1]$ \\
			&&&Output layer & 2$\times$ Tanh \\
			\bottomrule[1pt]
		\end{tabular}
	}
\end{table}

\subsection{The Implementation Details of Experiments}

\textbf{Experimental Datasets.} \tx{We employ Gym-Mujoco tasks to test the performance of DOMAIN. The Gym-Mujoco tasks, such as Halfcheetah-v2, Hopper-v2, and Walker2d-v2, are included in the D4RL benchmark \cite{ref29}, which are widely used in offline RL.} Fig. \ref{fig_2} depicts the above tasks. The goal is to control the robot's various joints to achieve faster and more stable locomotion. \cite{ref30}.

\tx{In our paper, we select $9$ datasets covering $3$ tasks. For three Gym-Mujoco tasks, three types of recorded datasets are considered: Medium, Medium-replay, and Medium-expert. The ``Medium'' dataset is generated by collecting 1 million samples from a policy that undergoes early stopping during online training using SAC \cite{ref21}. The ``Medium-replay'' dataset records all samples during the training process, reaching a medium level. The ``Medium-expert'' dataset is a combination of expert and sub-optimal data with a total 2 million samples \cite{ref30}}.

\textbf{Practical algorithm implementation details.} Fig. \ref{fig_3} gives the network composition of the algorithm, including environment, discriminator, critic, and policy network. \tx{Through the discriminator network, the model data sampling probability is calculated for the critic network update; the critic network guides the policy network update; the actions generated on the strategy network generate more model data through the environment model; the initial state $s$ is randomly sampled from the offline dataset and used to the generation of model data.} Table \ref{tab_1} provides the hyperparameters for the network model. The detailed parameter settings are given below:
\begin{figure*}
	\centering
	\includegraphics[width=0.95\textwidth]{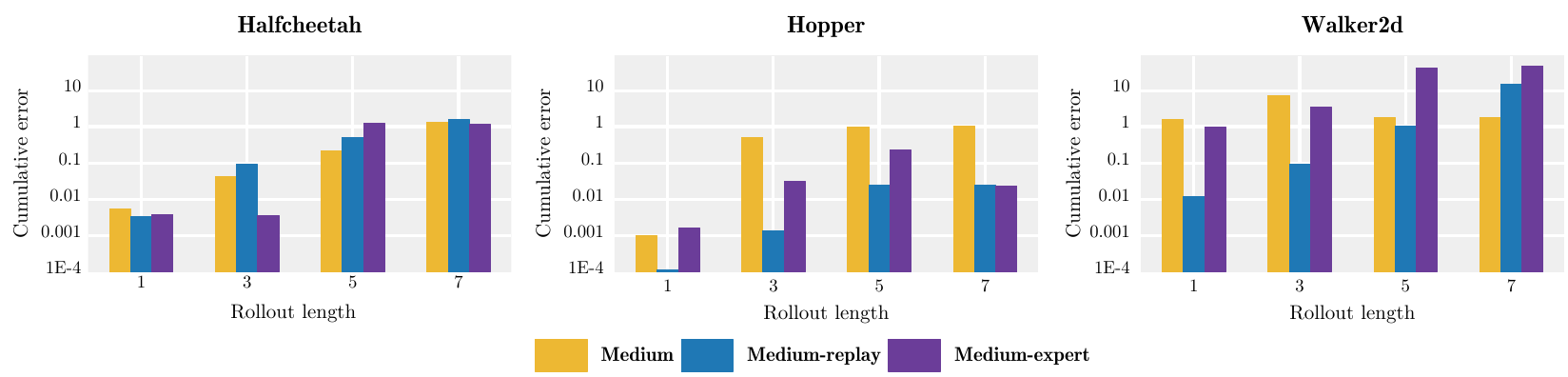}
	\caption{Quantitative analysis of the cumulative model errors. We compare the trained model with the actual MuJoCo environment, with the latter remaining unknown throughout the policy learning process. The initial state is initialized randomly within the MuJoCo environment, and the rollout policy is based on the final learned policy. To facilitate visualization, the logarithmic transformation is employed on the accumulated errors.}
	\label{fig4}
\end{figure*}

\begin{figure}
	\centering
	\includegraphics[width=6.5cm]{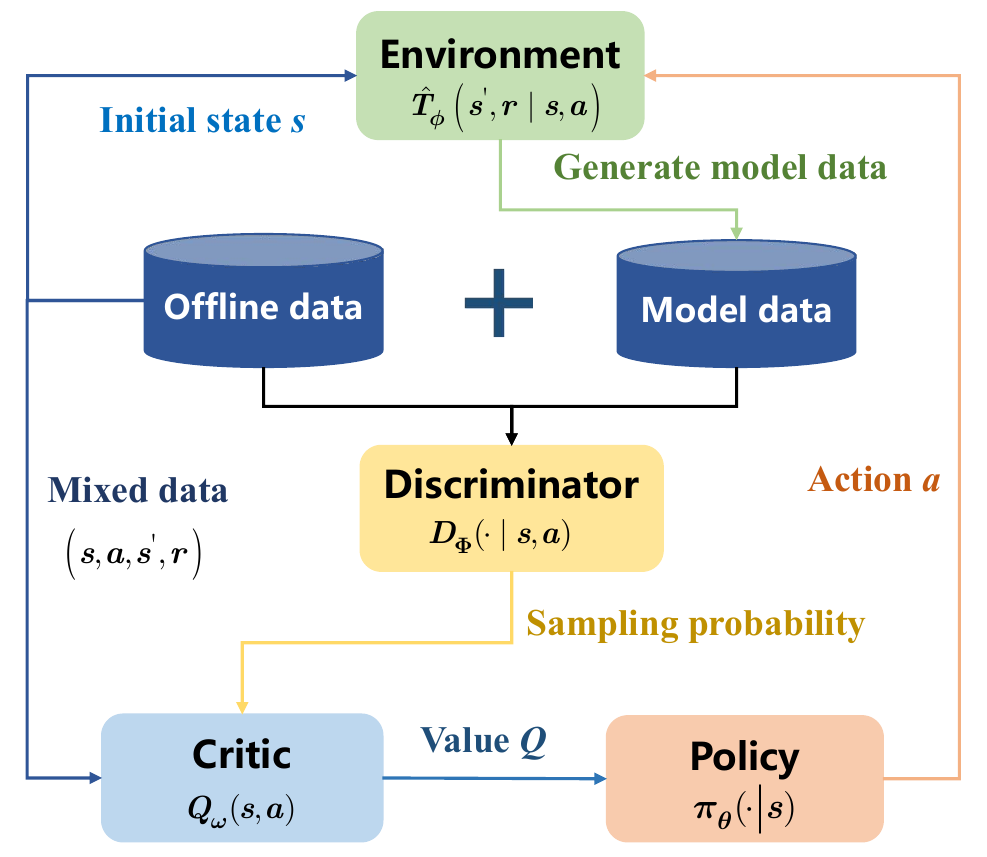}
	\caption{Network composition of the DOMAIN algorithm. The network is composed of four parts: environment, discriminator, critic, and policy.}
	\label{fig_3}
\end{figure}

1) \emph{Model Training}: The environment model is approximated by a probabilistic network that outputs a Gaussian distribution to obtain the next state and reward. We set $N$ as 7 and select the best 5 models. Each model consists of a 4-layer feedforward network with $200$ hidden units. The model employs maximum likelihood estimation with a learning rate of $1e-4$ and Adam optimizer. The $256$ data samples are randomly drawn from $\mathcal{D}_{\text {offline }}$ to train the model.

2) \emph{Discriminator Training}: Two discriminator networks, $D_{\Phi_{s a s}}\left(\cdot \mid s, a, s^{\prime}\right)$ and $D_{\Phi_{s a}}\left(\cdot \mid s, a\right)$, are trained to approximate the probabilities $p\left(\cdot \mid s, a, s^{\prime}\right), p\left(\cdot \mid s, a\right)$. The networks use the prediction ``$2 \times$ Tanh'' activation function to map the output values to the range $[-2,2]$, and then pass the clipped results through a softmax layer to obtain the final probabilities.

3) \emph{Policy Optimization}: Policy optimization is based on the SAC framework. The critic network $Q_{\omega}$ and the policy network $\pi_{\theta}$ adopt a 2-layer feedforward network with $256$ hidden units. The learning rates for them are set to $3e-4$ and $1e-4$ respectively. The ratio of model data is $f=0.5$. The entropy regularization coefficient $\alpha$ in Eq. \eqref{eq6} is automatically adjusted, with the entropy target set to -dim(A), and the learning rate for the self-tuning network is set to $1e-4$.

\begin{figure*}
	\centering
	\includegraphics[width=0.95\textwidth]{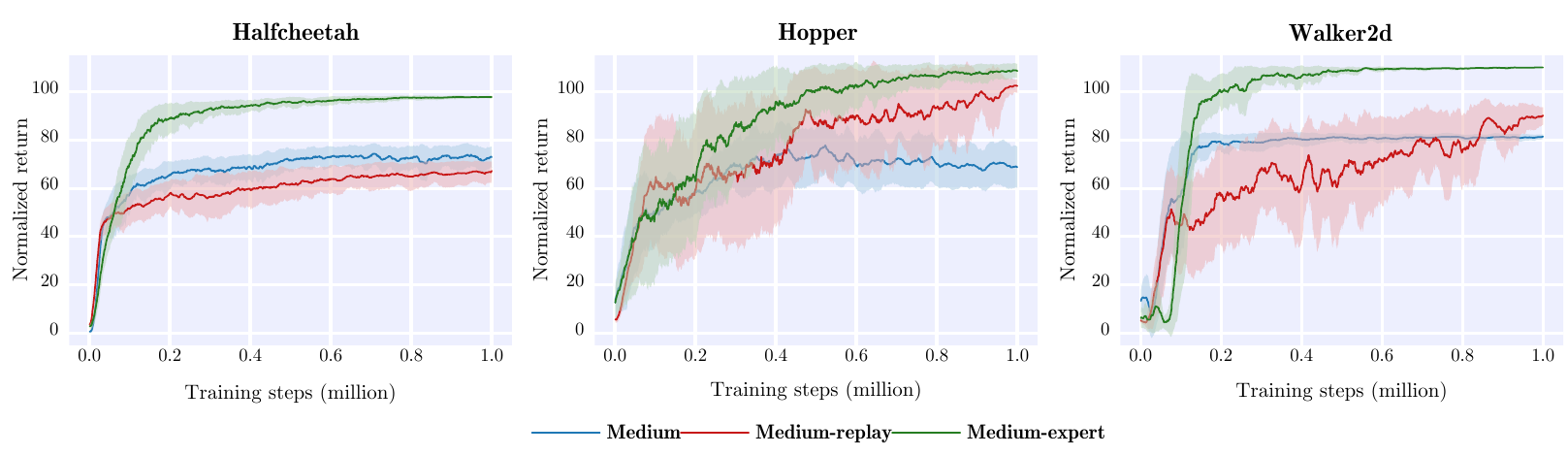}
	\caption{ Corresponding learning curves for Table \ref{tab_2}. Each figure shows the training curve for a specific task under different datasets, where the shaded area represents the standard variance of the return.}
	\label{fig7}
\end{figure*}

In practical policy optimization, the obtained adaptive sampling distribution is utilized to select model data with high uncertainty and penalize their $Q$-values. Following \cite{ref10,ref17}, we sample 10 actions in every high uncertainty state based on the uniform policy $Unif(a)$ and the learned policy $\pi(a \mid s)$, and then estimate the log-sum-exp term in \eqref{eq9}  using importance sampling.

\textbf{Parameter Tuning.} The quality of model data is influenced by the environment model $\widehat{T}_{\phi}$, the rollout policy $\mu(\cdot \mid s)$, and the rollout horizon $H$. The weight coefficient $\lambda$ of the regularization term significantly affects the stability of agent training. The rollout policy is the current learned policy. \tx{The parameters $H$ and $\lambda$ are two critical hyperparameters that require meticulous tuning to ensure optimal performance. The selection of the parameter $H$ is closely tied to the accuracy of the training environment model, and the regularization approach employed by the algorithm to handle erroneous model data. The choice of the parameter $\lambda$ is determined by the inherent distributional characteristics of the dataset. The detailed parameter analysis can be found in section \ref{Parameter Analysis}.} 

\tx{Moreover, note that RL is characterized by high training costs and instability, with the ideal approach being the identification of a set of hyperparameters that generalize well across various tasks, rather than requiring frequent tuning for each task. In practical applications, the computational cost of exhaustive parameter search is prohibitively high. This study is based on the COMBO framework, where the selection of parameters has two options. To maintain experimental rigor and comparative fairness, the set of parameter candidates is the same as COMBO. Table \ref{tab_5} gives the results of Gym-Mujoco tasks under different parameters. Based on this table, parameters $H$ and $\lambda$ for different tasks are set as: }

1) \emph{Rollout horizon $H$} : The rollout horizon $H$ is chosen from the set $\{1,5\}$. We uses $H=1$ for Walker2d-v2 tasks, and $H=5$ for Hopper-v2 and Halfcheetah-v2.

2) \emph{Weight coefficient $\lambda$ }: The parameter $\lambda$ reflects different strengths of regularization. The parameter $\lambda$ is chosen from the set $\{0.5,5\}$. For three medium-replay datasets and halfcheetah-medium datasets, $\lambda$ is chosen as 0.5, while for the remaining datasets, $\lambda$ is chosen as 5.

\begin{table*}
	\small
	\renewcommand{\arraystretch}{1.2}
	\caption{Results for D4RL tasks during the last 5 iterations of training averaged over 5 seeds. $\pm$ captures the standard deviation. Bold indicates the highest score. Highlighted numbers indicate results within 2\% of the most performing algorithm.}
	\label{tab_2}
	\centering
	\begin{tabular}{p{0.05cm}p{0.05cm}c|c|ccccc|ccc}
		\toprule[1pt] 
		\multicolumn{2}{c}{\multirow{2}{*}{\textbf{Type}}} &\multirow{2}{*}{\textbf{Environment}} &Ours & \multicolumn{5}{c|}{Model-based baseline} & \multicolumn{3}{c}{Model-free baseline} \\
		&& & \textbf{DOMAIN} & \textbf{ARMOR} & \textbf{RAMBO}&\textbf{COMBO} & \textbf{MOReL} & \textbf{MAPLE} & \textbf{MOAC} & \textbf{IQL} & \textbf{GORL}  \\
		\midrule
		\multicolumn{2}{c}{\multirow{3}{*}{\rotatebox{90}{Medium}}} & Halfcheetah & $72.7\pm4.7$ & $54.2$&\colorbox{blue!10}{$\mathbf{77.6}$} & $54.2$ & $42.1$ & $63.5$& $54.3$ & $47.4$ & $51.7$  \\
		&& Hopper & $70.3\pm8.9$ & \colorbox{blue!10}{$\mathbf{101.4}$} & $92.8$ & $97.2$& $95.4$ & $21.1$& $83.4$ & $66.3$ & $64.2$  \\
		&& Walker2d & $81.5\pm1.0$ & \colorbox{blue!10}{$\mathbf{90.7}$} & $86.9$ & $81.9$ & $77.8$ & $56.3$& $86.7$ & $78.3$ & $83.5$  \\
		\midrule
		\multirow{3}{*}{ \rotatebox{90}{Medium} } &\multirow{3}{*}{ \rotatebox{90}{Replay} }& Halfcheetah & \colorbox{blue!10}{$67.8\pm3.9$} &$50.5$& \colorbox{blue!10}{$\mathbf{68.9}$} & $55.1$ & $40.2$ & $50.3$ & $46.5$ & $45.4$ & $59.0$ \\
		&& Hopper & \colorbox{blue!10}{$\mathbf{102.7\pm2.3}$} &$97.1$& $96.6$ & $89.5$  & $93.6$ & $87.5$ & $98.0$ & $94.7$ & $74.7$  \\
		&& Walker2d & \colorbox{blue!10}{$\mathbf{90.4\pm3.6}$} &$85.6$& $85.0$ & $56.0$  & $49.8$ & $76.7$& \colorbox{blue!10}{$90.1$} & $73.9$ & $79.3$  \\
		\midrule
		\multirow{3}{*}{ \rotatebox{90}{Medium}} &\multirow{3}{*}{ \rotatebox{90}{Expert}} & Halfcheetah & \colorbox{blue!10}{$\mathbf{97.7\pm0.7}$} &$93.5$& $93.7$ & $90.0$ & $53.3$ &$63.5$& $87.2$ & $86.7$ & $88.2$  \\
		&& Hopper & \colorbox{blue!10}{$109.7\pm2.2$} &$103.4$& $83.3$ & \colorbox{blue!10}{${111.1}$} & \colorbox{blue!10}{$108.7$} & $42.5$ & \colorbox{blue!10}{{$\mathbf{111.4}$}} & $91.5$ & $88.8$  \\
		&& Walker2d &\colorbox{blue!10}{$110.1\pm0.1$}& \colorbox{blue!5}{$\mathbf{112.2}$}&$68.3$ & $103.3$ & $95.6$ & $73.8$ & $102.1$ & \colorbox{blue!10}{$109.6$} & \colorbox{blue!10}{$109.6$}  \\
		\midrule
		\multicolumn{3}{c|}{Average Score} & $\mathbf{89.2\pm3.0}$ & $87.6$&$83.7$ & $82.0$ & $72.9$ & $60.4$ & $84.8$ & $77.0$ & $76.3$ \\
		\bottomrule[1pt] 
	\end{tabular}
\end{table*}
\begin{table}
	\normalsize
	\caption{The results of D4RL task for different parameters. Bolded and highlighted numbers indicate the best performance.}
	\label{tab_5}
	\centering
	\resizebox{9cm}{!}{
		\begin{tabular}{p{0.02cm}p{0.02cm}c|cc|cc}
			\toprule[1pt] 
			\multicolumn{2}{c}{\multirow{2}{*}{\textbf{Type}}} &\multirow{2}{*}{\textbf{Environment}} & \multicolumn{2}{c|}{$\lambda=0.5$} & \multicolumn{2}{c}{$\lambda=5$} \\
			&& & $H=1$ &$H=5$& $H=1$ & $H=5$\\
			\midrule
			\multicolumn{2}{c}{\multirow{3}{*}{\rotatebox{90}{Medium}}} & Halfcheetah & $68.49$ & \colorbox{blue!10}{$\mathbf{72.69}$} & $51.12$ & $49.08$\\
			&& Hopper & $0.85$ & $69.94$ & $64.62$ &\colorbox{blue!10}{$\mathbf{70.29}$}\\
			&& Walker2d & $37.44$ & $78.57$ & \colorbox{blue!10}{$\mathbf{81.46}$} & $-0.10$ \\
			\midrule
			\multirow{3}{*}{ \rotatebox{90}{Medium} } &\multirow{3}{*}{ \rotatebox{90}{Replay} }& Halfcheetah & $54.40$& \colorbox{blue!10}{$\mathbf{67.75}$} & $50.00$ & $46.18$\\
			&& Hopper & $84.35$ &  \colorbox{blue!10}{$\mathbf{102.68}$} & $90.53$ & $100.6$ \\
			&& Walker2d & \colorbox{blue!10}{$\mathbf{90.36}$} & $85.08$ & $87.35$ & $52.69$\\
			\midrule
			\multirow{3}{*}{ \rotatebox{90}{Medium}} &\multirow{3}{*}{ \rotatebox{90}{Expert}} & Halfcheetah & $33.24$ & $50.31$ & $87.82$ & \colorbox{blue!10}{$\mathbf{97.74}$} \\
			&& Hopper & $1.84$ & $55.50$  & $101.1$ & \colorbox{blue!10}{$\mathbf{109.67}$}\\
			&& Walker2d &$10.72$ & $105.78$ &\colorbox{blue!5}{$\mathbf{110.12}$}& $109.02$ \\
			\bottomrule[1pt]
		\end{tabular}
	}
\end{table} 

\subsection{Experimental Results}
\textbf{Results on environment model prediction.} \tx{This section first analyzes and visualizes the accuracy of the trained environment model prediction before answering the above questions. Fig. \ref{fig4} gives the prediction results of cumulative error in rollout lengths 1,3,5, and 7 for different tasks and datasets, respectively.} In this figure, the predicted errors of four distinct datasets for tasks Halfcheetah, Hoppe, and Walker2d are arranged from left to right. The prediction error of the trained environmental model exhibits an increasing trend with the increment of steps across all datasets. Since the model exhibits the largest prediction error of the three Walker2d tasks, the rollout length is set to 1 in the Walker2d task to ensure the agent's stability.

However, single-step prediction is not conducive to the expansion of the OOD data by the environment model, because most of the predicted state-action data still falls within the region. As the rollout length $H$ increases, it is beneficial to explore the OOD region. Meanwhile, it is necessary to ensure the model data error is not too large. In Fig. \ref{fig4}, the prediction error of $H$ from 1 to 7 of the environmental model does not change too much under Halfcheetah and Hopper tasks. Within the error range, the optimal $H$ value should be determined by more experiments.

\textbf{Results on the D4RL tasks.}  This section compares recent model-free algorithms, such as MOAC \cite{ref51}, IQL \cite{ref31}, and GORL \cite{ref52}, as well as model-based algorithms, including ARMOR \cite{ref34}, MAPLE \cite{ref50}, MOReL \cite{ref16}, COMBO \cite{ref17}, and RAMBO \cite{ref20}. \tx{The results of the above methods are all based on the D4RL benchmark. The performance of these algorithms is evaluated based on the cumulative return in the robot motion tasks within the MuJoCo environment.} A higher score indicates better performance in terms of more stable and faster locomotion control. For comparison, the scores are normalized between 0 (random policy score) and 100 (expert policy score) \cite{ref30}. The normalized score $\tilde{S}$ is computed by:
	\begin{equation*}
		\tilde{S}=\frac{S-S_r}{S_e-S_r} \times 100.
	\end{equation*}
where $S_r$, $S_e$ and $S$ are the expected return of a random policy, an expert policy, and the trained policy by offline RL algorithms, respectively. The random policy scores under Halfcheetah, Hopper, and Walker2d are $-280.179$, $1.629$, and $-20.272$ respectively. The expert policy scores under Halfcheetah, Hopper, and Walker2d are $12135.0$, $4592.3$, and $3234.3$ respectively.

Table \ref{tab_2} presents the normalized scores of different methods. The reported scores are the normalized scores of the learned policies during the last 5 iterations, averaged over 5 random seeds.
The results for all methods are obtained from the original papers.
\tx{Due to the differences between datasets under different tasks, RL methods hardly perform well on all datasets. This table shows that our method DOMAIN performs poorly on the ``Medium'' dataset, ARMOR \cite{ref34} performs poorly on the ``Medium-replay'' dataset, and RAMBO \cite{ref20} performs poorly on the ``Medium-expert'' dataset.}
This table also indicates that the DOMAIN method achieves the best performance in three out of the nine datasets, and comparable results in three out of the remaining six settings. and a slightly poor performance in the three medium settings. Overall, DOMAIN can achieve better average performance across multiple datasets. Corresponding learning curves are shown in Fig. \ref{fig7}.

\tx{The reason why DOMAIN performs poorly in ``Medium'' dataset is that the distribution of medium is narrower than that of medium-replay and medium-expert. Then, the model data generated by the environment model trained based on the dataset is also narrow, and the trained discriminator hardly distinguishes model data from the offline dataset. As a result, the adaptive sampling distribution $\omega(s,a)$ is inaccurate, which reduces the final agent performance. Moreover, Table \ref{tab_2} shows that the performance of the ``Medium-expert'' dataset in three tasks is almost always superior to that of ``Medium'' and ``Medium-replay'' datasets for any algorithms. The difference between ``Medium'' and ``Medium-replay'' is that ``Medium-replay'' contains more data at different levels, which can facilitate agent exploration. However, the algorithm may not necessarily perform better than ``Medium'' under ``Medium-replay'' for different tasks. This mainly depends on the adaptability of the algorithm itself to diverse data and the coverage of the data.}

\begin{table*}
	\small
	\renewcommand{\arraystretch}{1.4}
	\caption{The parameter configuration of our designed simple MDP.}
	\label{tab:4}
	\centering
	\begin{tabular}{cccccccccc}
		\toprule[1pt]  
		\multirow{2}{*}{\textbf{Parameter}}& \multicolumn{6}{c}{\textbf{Designed Dynamic Model}}&& \multicolumn{2}{c}{\textbf{Offline Dataset}}\\
		\cline{2-7}\cline{9-10}
		& $a$&$[-1,-0.6)$&$[-0.6,-0.2)$&$[-0.2,0.2)$&$[0.2,0.6)$&$[0.6,1]$&&Behavioral Policy & Data Size\\
		\cline{1-7}\cline{9-10}
		\multirow{2}{*}{\textbf{Value}} & $\mu$&$-a+0.2$&$5(a+0.4)^2+0.6$&$-5a^2+1$&$10(a-0.4)^2+0.4$&$a+0.2$&&\multirow{2}{*}{Random Policy}&\multirow{2}{*}{$1000$}\\
		&$\sigma$& $0.06$&$0.04$&$0.02$&$0.04$&$0.06$\\
		\bottomrule[1pt]
	\end{tabular}
\end{table*}

\begin{table} 
	\renewcommand{\arraystretch}{1.1}
     \caption{Comparison results of computation time (unit: seconds).}
    \label{tab_cost}
	\centering
	{
		\begin{tabular}{ccccc}
			\toprule[1pt]
			\multirow{2}{*}{\textbf{Method}}&\multicolumn{3}{c}{\textbf{Environment}}&\multirow{2}{*}{\textbf{Average}}\\
			\cmidrule(lr){2-4}
			&Walker2d&Hopper&Halfcheetah&\\
			\midrule
			{\textbf{COMBO}}&$ 44422.1$&$41477.9$&$49271.5$&$43908.6$\\
			{\textbf{RAMBO}}&$91853.9$&$86531.3$&$96766.3$&$91717.2$\\
            {\textbf{DOMAIN}}&$46957.7$&$46930.9$&$48244.5$&$47377.7$\\
			\toprule[1pt]
		\end{tabular}
	}
\end{table}

\tx{The computational complexity is an important index for the performance of the proposed method. The training time of model-based offline RL usually includes two parts: environment learning and policy learning. Table \ref{tab_cost} compares computation cost between DOMAIN and the current two typical model-based offline RL methods (COMBO \cite{ref17} and RAMBO \cite{ref20}) under the three environments. This table shows that DOMAIN can reduce computational costs by nearly half compared to RAMBO, primarily because the environment model is continuously updated during the policy learning process in RAMBO. The computational complexity of DOMAIN is slightly higher than that of COMBO, as DOMAIN requires the training of a discriminator to solve the adaptive sampling distribution $\omega$, which introduces additional training costs. In general, compared with the above two methods, DOMAIN can achieve better control performance with relatively small computational complexity.}

\begin{figure}
	\centering
	\includegraphics[width=9cm]{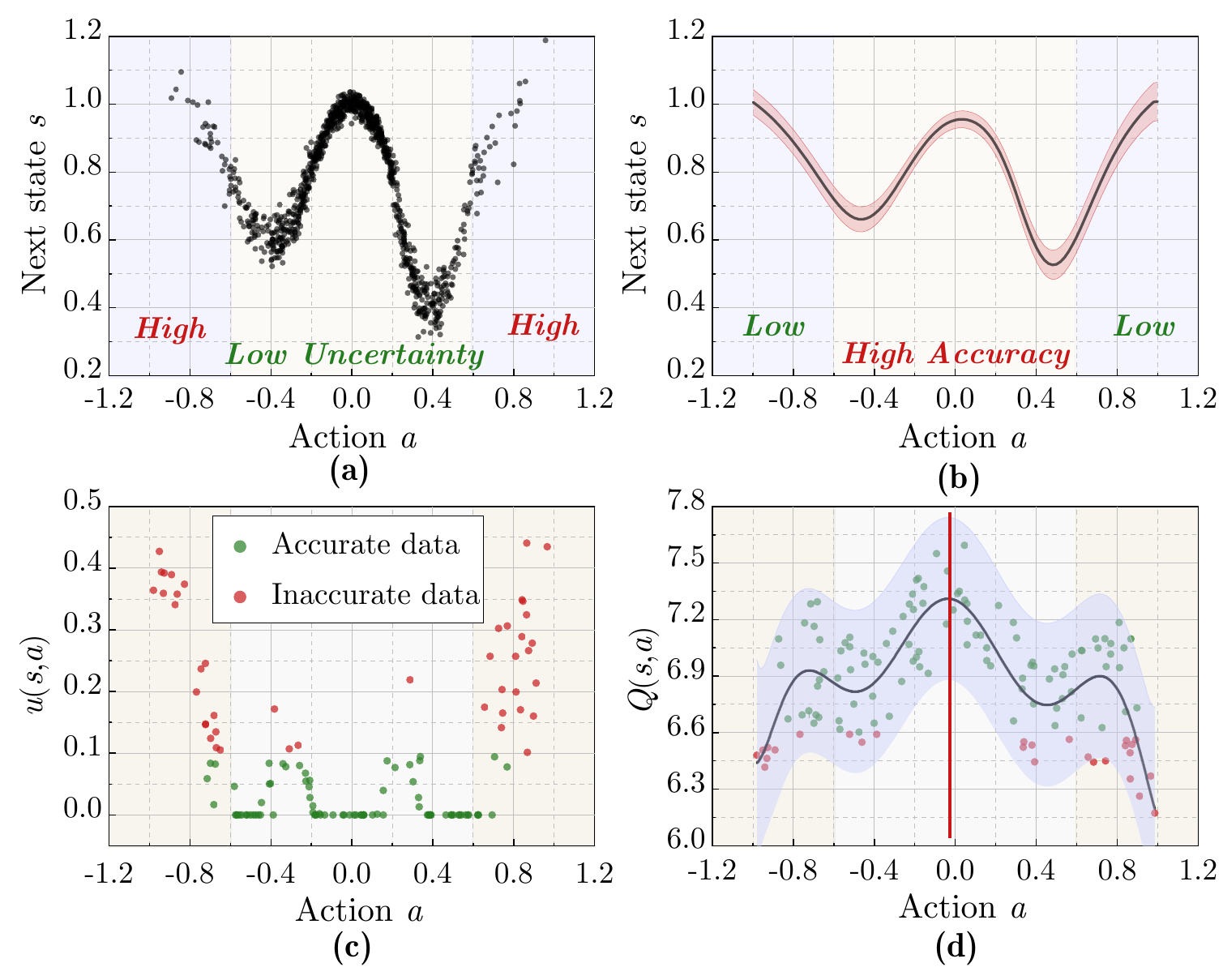}
	\caption{The visualization results of the DOMAIN algorithm on the designed MDP. (a) illustrates the distribution of collected offline data. The intermediate color region represents the low uncertainty area, while the region on both sides represents the high uncertainty area; (b) presents the prediction results of the trained dynamic model under the initial state ${s=0}$. The red area represents the range of variance predicted by ensemble models; (c) depicts the error distribution of model data. The green color represents precise model data, while the red color represents imprecise model data; (d) illustrates the distribution of $Q$-values of model data. The black curve is the high-order fitting curve of the data. The red line indicates the optimal action, i.e. ${a\approx 0}$.}
	\label{fig6}
\end{figure}

\subsection{ Visualization of DOMAIN Performance}
Due to the high dimension of states and actions in the D4RL tasks, visualizing the performance is greatly challenging. Therefore, a simple example MDP is defined for illustrative purposes following \cite{ref20}. \tx{For this MDP, both the state space $\mathcal{S}$ and action space $\mathcal{A}\in [-1,1]$ are one-dimensional. The reward is defined as $R(s,a)=s$, where a larger value of state $s$ leads to a higher reward.} The dynamic transition model is given by $T(s^{\prime}\mid s,a)=\mathcal{N}(\mu, \sigma)$ with the parameters shown in Table \ref{tab:4}, indicating that the next state depends on the current action, independent of the initial state. To visualize the DOMAIN exploration in the OOD region, we sample actions from $\mathcal{N}(\mu=0, \sigma=0.35)$, which means that the behavioral policy for the offline dataset is a random policy.

\tx{Fig. \ref{fig6} (a) shows the distribution of the collected offline data for the aforementioned MDP, illustrating that the actions do not cover all regions. Fig. \ref{fig6} (b) presents the prediction results of the trained dynamic model, which demonstrate higher accuracy in the in-distribution region (low-uncertainty area) and lower accuracy in the OOD region (high-uncertainty area).} Fig. \ref{fig6} (c) displays the model data error after 3000 iterations. It can be observed that the model data within the region exhibits relatively low errors, while the data outside the region shows larger errors. This result is consistent with the prediction results of the trained dynamic model. Fig. \ref{fig6} (d) shows the model data $Q$-value curve in state ${s=0}$ after 3000 iterations. We can find that the DOMAIN algorithm does not universally underestimate all model data. Instead, it is more prone to underestimating inaccurate model data. Moreover, the value of the optimal action, i.e.,$\arg \max _{a} Q(s,a)$, is approximately zero, which aligns with the selection of the offline dataset. This is because the reward is solely dependent on the state, and the state value is maximized when ${a\approx0}$ is in-distribution region. The consistency between model data and offline data in selecting the optimal actions contributes to enhancing the stability of agent training.

\subsection{\tx{Robust Experiments}}
\tx{The above experiments are based on a simulation environment. However, there is a gap between the simulation and the actual environment, and there are different types of disturbances in the practical scenario. Therefore, it is a necessity to test the performance of the proposed method in the above situations. Here, we conduct two types of experiments: one is the simulation environment parameter perturbation experiment, and another is the external disturbance experiment.}

\begin{figure}
	\centering
	\includegraphics[width=9cm]{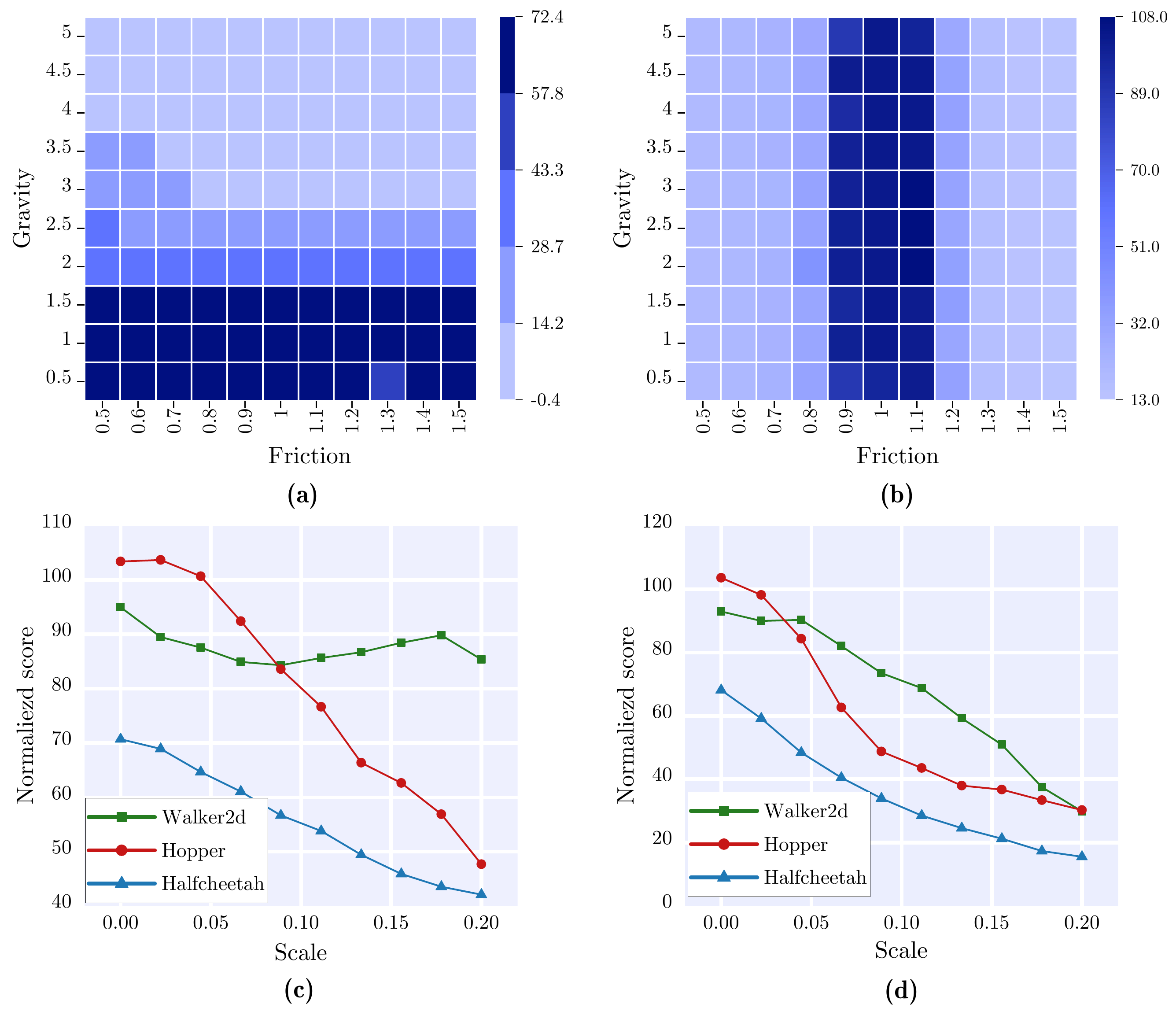}
	\caption{The results of robust experiments for DOMAIN. (a) and (b) give the performance under environment parameter random perturbations in halfcheetah and hopper environments, respectively.The gravity and friction vary from 0.5 to 5 times and 0.5 to 1.5 times the value of the simulation environment, respectively. (c) and (d) show the performance under attack scales range $[0, 0.2]$ of RA and AD two attack types in walker2d, hopper and halfcheetah environments, respectively}
	\label{fig88}
\end{figure}

\tx{\textbf{Environment parameter perturbations:} Achieving high-fidelity simulation environments is often challenging, which leads to inherent sim-to-real gaps in policy learning. In the case of D4RL tasks, the simulation environment relies on the MuJoCo physics engine. To assess the performance of our method in environments with large dynamics gaps, we modify key MuJoCo parameters, including friction and gravity, to create an environment with a substantial sim-to-real gap. Fig. \ref{fig88} (a) and (b) present the results under various test conditions, where different gravity and friction coefficients are applied to the environment, with darker colors indicating superior performance. This figure also illustrates that under halfcheetah, variations in gravity have a significant impact on performance. Conversely, under hopper, changes in friction exert a substantial influence on performance.} 

\tx{\textbf{External disturbances:} Due to the observation errors and external disturbances in real-world scenarios, the robustness of the policy is evaluated using the two perturbation methods below, referred to as RA and AD \cite{r_9,r_10}. For RA method, the perturbed states are randomly sampled from the set $U_{\epsilon}(s)=\{s'\in \mathcal{S}:d(s,s')^p\leq \epsilon\}$, where $\epsilon$ denotes the scale of perturbation and $d(s,s')^p$ measures the distance between $s$ and $s'$ \cite{r_9}. AD is an effective perturbation method based on the agent's policy, which selects the state $s'$ from $U_{\epsilon}(s)$ that maximizes the gap between action distribution $\pi_\theta(\cdot |s)$ and $\pi_\theta(\cdot |s')$, i.e. $\min_{s'\in U_\epsilon(s)}-D_J(\pi_\theta(\cdot |s)||\pi_\theta(\cdot |s'))$ \cite{r_7}. 
The observation states are attacked through the above perturbation in the evaluation phase. Fig. \ref{fig88} (c) and (d) give the result of DOMAIN with varying perturbation scale $\epsilon$ under RA and AD attack in three environments. As external disturbances progressively increase, the performance exhibits a declining trend.}  

\tx{Based on the experimental results presented in Fig. \ref{fig88}, it is evident that DOMAIN is capable of adapting to minor external disturbances and variations in environmental parameters. However, its performance significantly deteriorates under large-scale variations. Future research should focus on enhancing the robustness of the algorithm to address these limitations. For example, the robust Bellman operator can be introduced into the standard Bellman operator, or the idea of the robust adversarial model can be introduced into model training.}

\begin{table*}
	\small
	\renewcommand{\arraystretch}{1.2}
	\caption{\tx{The agent performance results averaged over 5 seeds under different parameters $H$ and $\lambda$. Bold indicates the highest score in ablation experiments. Highlighted numbers indicate the optimal results in Table \ref{tab_5}.}}
	\label{tab_44}
	\centering
	\begin{tabular}{p{0.04cm}p{0.04cm}c|ccccc|ccccc}
		\toprule[1pt] 
		\multicolumn{2}{c}{\multirow{2}{*}{\textbf{Type}}} &\multirow{2}{*}{\textbf{Environment}} & \multicolumn{5}{c|}{Analysis for weight $\lambda$} & \multicolumn{5}{c}{Analysis for rollout length $H$} \\
		&& & $H$ &$\lambda=0.1$ & $\lambda=0.5$ & $\lambda=1$&$\lambda=5$ & $\lambda$& $H=1$ & $H=3$ & $H=5$ & $H=7$ \\
		\midrule
		\multicolumn{2}{c}{\multirow{3}{*}{\rotatebox{90}{Medium}}} & Halfcheetah & $5$ & $70.1$&\colorbox{blue!10}{$72.7$} & $65.1$ & $49.1$ & $0.5$ & $68.5$ & $72.5$ &\colorbox{blue!10}{$72.7$}& $\mathbf{74.5}$ \\
		&& Hopper & $5$ & $13.6$ & $69.9$ & $24.3$& \colorbox{blue!10}{$70.3$} & $5.0$ & $64.6$ & $\mathbf{81.6}$ & \colorbox{blue!10}{$70.3$}&$68.1$ \\
		&& Walker2d & $1$ & $16.3$ & $37.4$ & $64.3$ & \colorbox{blue!10}{$\mathbf{81.5}$} & $5.0$ & \colorbox{blue!10}{$\mathbf{81.5}$} & $11.3$ & $46.2$ &$22.2$\\
		\midrule
		\multirow{3}{*}{ \rotatebox{90}{Medium} } &\multirow{3}{*}{ \rotatebox{90}{Replay} }& Halfcheetah & $5$&$\mathbf{68.3}$ &\colorbox{blue!10}{$67.8$}&$64.4$& $46.2$ & $0.5$ & $54.4$ & $65.9$ & \colorbox{blue!10}{$67.8$}& $65.0$ \\
		&& Hopper & $5$ &$13.8$& \colorbox{blue!10}{$102.7$} & $77.2$  & $100.6$ & $0.5$ & $84.4$ & $\mathbf{103.8}$ & \colorbox{blue!10}{$102.7$}&$103.7$ \\
		&& Walker2d & $1$ & $44.4$ &\colorbox{blue!10}{$\mathbf{90.4}$}& $86.0$ & $87.4$  & $0.5$ &\colorbox{blue!10}{$\mathbf{90.4}$}& $83.5$ & $85.1$ & $81.0$ \\
		\midrule
		\multirow{3}{*}{ \rotatebox{90}{Medium}} &\multirow{3}{*}{ \rotatebox{90}{Expert}} & Halfcheetah & $5$ &$56.7$& $50.3$ & $\mathbf{100.9}$ & \colorbox{blue!10}{$97.7$} & $5.0$ & $87.8$ & $98.0$ &\colorbox{blue!10}{$97.7$}&$97.6$ \\
		&& Hopper & $5$ &$10.2$& $55.5$ & $104.2$&\colorbox{blue!10}{$109.7$} & $5.0$ & $101.1$ & $109.3$ & \colorbox{blue!10}{$109.7$} & $\mathbf{111.2}$ \\
		&& Walker2d &$1$& $8.6$&$10.7$ & $0.9$ & \colorbox{blue!10}{$\mathbf{110.1}$} & $5.0$ &\colorbox{blue!10}{$\mathbf{110.1}$}& $103.5$ & $109.0$ & $83.0$\\
		\bottomrule[1pt]
	\end{tabular}
\end{table*}

\subsection{The Parameter Analysis} \label{Parameter Analysis}
This part gives further analysis for weight $\lambda$ and rollout length $H$. Note that the tuning results in Table \ref{tab:4} are not optimal with high probability. Table \ref{tab_44} analyzes the impact of two parameter changes on the performance of the agent. Due to high training cost, we fix one parameter and test the impact of changes in another parameter on agent performance. \tx{Note that the tuning results presented in Table \ref{tab_5} are not optimal with high probability. The optimal parameters can be found through more precise and wide-ranging parameter searches, but this requires significant training costs. The parameter ranges specified in Table \ref{tab_44} are primarily intended to analyze the effect of parameter variations on agent performance.} 

\textbf{The effect of weight $\lambda$.} The parameter $\lambda$ is used to regularization between Bellman error and $Q$ value of data, and guarantees the lower bound of $Q$-function for inaccuracy model data. \tx{A larger $\lambda$ is needed to guarantee the lower bound of inaccuracy model data in a narrower data distribution. However, for diverse data distribution, too large $\lambda$ may cause the lower bound to be too loose, thus reducing agent performance. Table \ref{tab_44} shows that in narrower ``Medium'' and ``Medium-Expert'' datasets, too small $\lambda$ leads to unstable training, thus reducing the agent's performance. For diverse ``Medium-Replay'' datasets, too large $\lambda$ causes the high constraint, thus limiting the improvement of the agent.}

\textbf{The effect of rollout length $H$.} When rollout length $H=1$, the trained environment model has high prediction accuracy. \tx{As $H$ increases, the accuracy of prediction decreases, but it is beneficial to explore the OOD region. If the model data error is too large, the algorithm will not be able to achieve a balance between accurate offline data and inaccurate model data, leading to unstable agent training. Therefore, it is necessary to ensure that the model data error is not too large.} Table \ref{tab_44} shows that, as $H$ increases, the agent's performance shows a downward trend under the Walker2d task since the error of the trained environment is large (Fig. \ref{fig4}). For Halfcheetah and Hopper tasks, the environment error is small. Therefore, as $H$ increases, the agent's performance does not show a downward trend, but the optimal $H$ is different for different datasets.

%% file: tex/05-Conclusion.tex
\section{Conclusion}\label{sect6}
This paper proposed a mildly conservative model-based offline RL (DOMAIN) algorithm. By designing a model data adaptive sampling distribution, DOMAIN can dynamically adjust the model data penalty, increasing the penalty for inaccuracy model data while reducing the penalty for accurate model data, thereby reducing its conservatism. Moreover, DOMAIN avoided unreliable estimation of model uncertainty. Theoretical analysis demonstrated that DOMAIN can guarantee safe policy improvements and $Q$-values learned by DOMAIN outside the region are lower bounds of the true $Q$-values. Experimental results on the D4RL benchmark showed that DOMAIN outperforms previous RL algorithms on most datasets. {Despite the promising results, this study presents several limitations that warrant further investigation:} 

{1) The incorporation of conservatism during the training process inherently restricts the potential performance improvement of the agent. Future work could explore the optimization of the environment model itself to achieve enhanced agent performance without compromising safety constraints.} 

{2) The quality of model-generated data is highly dependent on the rollout length, which was empirically determined for different tasks through manual tuning in this study. To improve the generalizability of the proposed method, subsequent research should investigate an adaptive rollout strategy that dynamically adjusts the rollout length based on real-time data quality metrics.} 

{3) The experimental validation of the proposed algorithm has been confined to simulation environments. To further demonstrate its practical applicability, future research should include offline data collection from physical robotic locomotion tasks, followed by comprehensive network training and policy deployment in real-world scenarios.}

%% file: tex/06-Appendix.tex
\newpage
\clearpage
\setcounter{page}{1}

\onecolumn
\begin{center}
    \LARGE\textbf{Supplementary Materials (Appendix)} \\
    \vspace{0.5cm}
\end{center}

\setcounter{equation}{0}
\renewcommand{\theequation}{S.\arabic{equation}}
\setcounter{figure}{0}
\renewcommand{\thefigure}{S.\arabic{figure}}
\setcounter{table}{0}
\renewcommand{\thetable}{S.\arabic{table}}
\tx{This section provides the detailed proof for the convergence of Bellman operator, the lower bound for value function, mildly conservative estimation of $Q$-value and policy improvement guarantee (Theorem \ref{thm4}- \ref{thm3}).}

\subsection{Proof of Theorem \ref{thm4}}\label{proof4}
Before proof, we provide a notation clarification. Let $\mathcal{T}^{\pi}$ denote the ideal Bellman operator, and $\widehat{\mathcal{T}}^{\pi}$ represents the actual Bellman operator. $\mathcal{M}$ refers to the true MDP, $\widehat{\mathcal{M}}$ represents the learned MDP model, and $\overline{\mathcal{M}}$ denotes the empirical MDP, where ``empirical'' refers to operations based on samples. $\mathcal{T}_{\mathcal{M}}^{\pi},  T_{\mathcal{M}}$, and $r_{\mathcal{M}}$ denote the Bellman operator, state transition, and reward under the true MDP $\mathcal{M}$, respectively. $\mathcal{T}_{\widehat{\mathcal{M}}}^{\pi},  T_{\widehat{\mathcal{M}}}$, and $r_{\widehat{\mathcal{M}}}$ represent the corresponding operators under the learned MDP $\widehat{\mathcal{M}}$, while $\mathcal{T}_{\overline{\mathcal{M}}}^{\pi}, T_{\overline{\mathcal{M}}}$ and $r_{\overline{\mathcal{M}}}$ denote the operators under the empirical MDP $\overline{\mathcal{M}}$. Note that $\mathcal{T}^{\pi}=\mathcal{T}_{\mathcal{M}}^{\pi}$ and $\widehat{\mathcal{T}}^{\pi}=(1-f) \mathcal{T}_{\overline{\mathcal{M}}}^{\pi}+f \mathcal{T}_{\widehat{\mathcal{M}}}^{\pi}$. 

{\textit{\textbf{Proof.}}} ${\mathcal{T}}_{\overline{\mathcal{M}}}^{\pi} Q(s, a)$ is empirical Bellman operator, denoted as ${\mathcal{T}}_{\overline{\mathcal{M}}}^{\pi} Q(s, a) = r(s, a)+\gamma\max_{a'\in \mathcal{A}}Q\left(s^{\prime}, a^{\prime}\right)$. $\mathcal{T}_{\widehat{\mathcal{M}}}^{\pi}$ is Bellman operator under learned MDP $\widehat{\mathcal{M}}$, denoted as $\mathcal{T}_{\widehat{\mathcal{M}}}^{\pi} = r(s,a)+\gamma\mathbb{E}_{s'\sim T_{\widehat{\mathcal{M}}}}[\max_{a'\in \mathcal{A}}Q(s',a')]$. Then, according to $
\widehat{\mathcal{T}}^{\pi}=(1-f) \mathcal{T}_{\overline{\mathcal{M}}}^{\pi}+f \mathcal{T}_{\widehat{\mathcal{M}}}^{\pi}
$, we can derive:
\begin{equation}
\begin{aligned}
\widehat{\mathcal{T}}^{\pi}Q(s,a)&=(1-f) \left\{r(s, a)+\gamma\max_{a'\in \mathcal{A}} Q\left(s^{\prime}, a^{\prime}\right)\right\}+f \left\{r(s,a)+\gamma\mathbb{E}_{s'\sim T_{\widehat{\mathcal{M}}}}[\max_{a'\in \mathcal{A}}Q(s',a')]\right\}
\end{aligned} 
\end{equation}

Let $Q_1$ and $Q_2$ be two arbitrary $Q$-function, and $\|\cdot\|_\infty$ is $\mathcal{L}_\infty$ norm, then we can get the following equation:
\begin{equation}
\begin{aligned}
&\|\widehat{\mathcal{T}}^\pi Q_1-\widehat{\mathcal{T}}^\pi Q_2\|_{\infty}\\ 
=& (1-f)\gamma\max_{s,a}\left|\max_{a'\in \mathcal{A}} Q_1\left(s^{\prime}, a^{\prime}\right)-\max_{a'\in \mathcal{A}} Q_2\left(s^{\prime}, a^{\prime}\right)\right|+ f\gamma\max_{s,a}\left|\mathbb{E}_{s'\sim T_{\widehat{\mathcal{M}}}}[\max_{a'\in \mathcal{A}}Q_1]-\mathbb{E}_{s'\sim T_{\widehat{\mathcal{M}}}}[\max_{a'\in \mathcal{A}}Q_2]\right|\\
\leq& (1-f)\gamma\max_{s,a}\left|\max_{a'\in \mathcal{A}} Q_1\left(s^{\prime}, a^{\prime}\right)-\max_{a'\in \mathcal{A}} Q_2\left(s^{\prime}, a^{\prime}\right)\right|+ f\gamma\max_{s,a}\mathbb{E}_{s'\sim T_{\widehat{\mathcal{M}}}}\left|[\max_{a'\in \mathcal{A}}Q_1-\max_{a'\in \mathcal{A}}Q_2]\right|\\
\leq& (1-f)\gamma\left\| Q_1- Q_2\right\|_{\infty}+ f\gamma\left\| Q_1- Q_2\right\|_{\infty}\\
=& \gamma\left\| Q_1- Q_2\right\|_{\infty},\qquad\text{where } \gamma\in(0,1)
\end{aligned} 
\end{equation}

According to the fixed point theorem, any initial $Q$ function can converge to a unique fixed point $Q^*$ by repeatedly applying the Bellman operator $\widehat{\mathcal{T}}^\pi$. \tx{This completes the proof of Theorem \ref{thm4}.}

\subsection{Proof of Theorem \ref{thm1}}\label{proof1}
Before proving Theorem \ref{thm1}, we make the following assumptions following prior works \cite{ref10,ref17,ref24,ref28}:
\newtheorem{assumption}{\bf Assumption}
\begin{assumption}\label{as1}
For $\forall s, a \in \mathcal{D}$, the following inequality holds with a probability greater than $1-\delta$, where $\delta \in(0,1)$:
\begin{equation}\label{p-eq1}
    \begin{aligned}
        &\left|r_{\mathcal{M}}(s, a)-r_{\overline{\mathcal{M}}}(s, a)\right|<\frac{C_{r, \delta}}{\sqrt{|\mathcal{D}(s, a)|}}, \qquad \left|T_{\mathcal{M}}\left(s^{\prime} \mid s, a\right)-T_{\overline{\mathcal{M}}}\left(s^{\prime} \mid s, a\right)\right|_{1}<\frac{C_{T, \delta}}{\sqrt{|\mathcal{D}(s, a)|}}
    \end{aligned}
\end{equation}
where $|\mathcal{D}(s, a)|$ represents the cardinality (number of occurrences) of a specific state-action pair $(s, a)$ in the dataset $\mathcal{D}$. For $(s, a) \notin \mathcal{D}$, we assume $|\mathcal{D}(s, a)|$ to be less than 1. The reward is bounded within a certain range, given by $\left|r_{\mathcal{M}}(s, a)\right|<R_{\max }$. 
\end{assumption} 

\newtheorem{lemma}{\bf Lemma}
\begin{lemma}\label{le1}
For any policy $\pi$, the disparity between the actual Bellman operator $\mathcal{T}_{\mathcal{M}}^{\pi}$ and the Bellman operators $\mathcal{T}_{\overline{\mathcal{M}}}^{\pi}$ under the empirical MDP, as well as the disparity between the actual Bellman operator $\mathcal{T}_{\mathcal{M}}^{\pi}$ and the Bellman operators $\mathcal{T}_{\widehat{\mathcal{M}}}^{\pi}$ under the learned MDP satisfy the following inequalities with high probability $1-\delta$, respectively.
    \begin{equation*}\label{p-eq2}
        \begin{aligned}
            &\left|\mathcal{T}_{\mathcal{M}}^{\pi} \widehat{Q}^{k}-\mathcal{T}_{\overline{\mathcal{M}}}^{\pi} \widehat{Q}^{k}\right|<\frac{C_{r, T, \delta} R_{\max }}{(1-\gamma) \sqrt{|\mathcal{D}(s, a)|}}, \qquad \left|\mathcal{T}_{\mathcal{M}}^{\pi} \widehat{Q}^{k}-\mathcal{T}_{\widehat{\mathcal{M}}}^{\pi} \widehat{Q}^{k}\right|<\varepsilon_{r}+D\left(T_{\mathcal{M}}, T_{\widehat{\mathcal{M}}}\right) \frac{2 \gamma R_{\max }}{1-\gamma}
        \end{aligned}
    \end{equation*}
\end{lemma}

\tx{\textbf{Proof for Lemma \ref{le1}.}}  The difference between the Bellman operator $\mathcal{T}_{\widehat{\mathcal{M}}}^{\pi}$ under the empirical MDP and the actual Bellman operator $\mathcal{T}_{\mathcal{M}}^{\pi}$ is given by:
\begin{equation}\label{p-eq3}
\begin{aligned}
    \left|\mathcal{T}_{\mathcal{M}}^{\pi} \widehat{Q}^{k}-\mathcal{T}_{\overline{\mathcal{M}}}^{\pi} \widehat{Q}^{k}\right|\leq&\Big|r_{\mathcal{M}}(s, a)-r_{\overline{\mathcal{M}}}(s, a)\Big|+\gamma\Big|\sum_{s^{\prime}}(T_{\mathcal{M}}\left(s^{\prime} \mid s, a\right)-T_{\overline{\mathcal{M}}}\left(s^{\prime} \mid s, a\right)) \mathbb{E}_{a^{\prime} \sim \pi}\left[\widehat{Q}^{k}\left(s^{\prime}, a^{\prime}\right)\right]\Big| \\
    \leq & \frac{(1-\gamma) C_{r, \delta}+\gamma C_{T, \delta} 2 R_{\max }}{(1-\gamma) \sqrt{|\mathcal{D}(s, a)|}}=\frac{C_{r, T, \delta} R_{\max }}{(1-\gamma) \sqrt{|\mathcal{D}(s, a)|}}
\end{aligned}
\end{equation}

Here, $C_{r, T, \delta}$ is a function representing a combination of $C_{r, \delta}$ and $C_{T, \delta}$. Similarly, the difference between the Bellman operator $\mathcal{T}_{\widehat{\mathcal{M}}}^{\pi}$ under the learned MDP and the actual Bellman operator $\mathcal{T}_{\mathcal{M}}^{\pi}$ can be derived as:
\begin{equation}\label{p-eq4}
\begin{aligned}
    \left|\mathcal{T}_{\mathcal{M}}^{\pi} \widehat{Q}^{k}-\mathcal{T}_{\widehat{\mathcal{M}}}^{\pi} \widehat{Q}^{k}\right|\leq & \Big|r_{\mathcal{M}}(s, a)-r_{\widehat{\mathcal{M}}}(s, a)\Big|+\gamma\Big|\sum_{s^{\prime}}(T_{\mathcal{M}}\left(s^{\prime} \mid s, a\right)-T_{\widehat{\mathcal{M}}}\left(s^{\prime} \mid s, a\right)) \mathbb{E}_{a^{\prime} \sim \pi}\left[\widehat{Q}^{k}\left(s^{\prime}, a^{\prime}\right)\right]\Big| \\
    \leq & \Big|r_{\mathcal{M}}(s, a)-r_{\widehat{\mathcal{M}}}(s, a)\Big|+D\left(T_{\mathcal{M}}, T_{\widehat{\mathcal{M}}}\right) \frac{2 \gamma R_{\max }}{1-\gamma}
\end{aligned}
\end{equation}
where $\varepsilon_{r}(s, a)$ represents the error between the learned reward and the actual reward, and $D\left(T_{\mathcal{M}}, T_{\widehat{\mathcal{M}}}\right)$ denotes the discrepancy between the transitions of the learned model and the true model. Lemma \ref{le1} is thus proven.

\tx{\textbf{Proof for Theorem \ref{thm1}.}} First, we introduce the following notation: Let $\mathcal{T}^{\pi} Q=r+\gamma P^{\pi} Q$, where $P^{\pi} Q(s, a)=\mathbb{E}_{s^{\prime} \sim T(\cdot\mid s, a), a^{\prime} \sim \pi(\cdot \mid s^{\prime})}[Q^{\pi}(s^{\prime}, a^{\prime})]$. Based on Eq. \eqref{eq14} and the Lemma \ref{le1}, it can obtain:
\begin{equation}\label{p-eq5}
\begin{aligned}
    & \widehat{Q}^{k+1}=(1-f) \mathcal{T}_{\overline{\mathcal{M}}}^{{\pi}} \widehat{Q}^{k}+f \widehat{\mathcal{T}}_{\widehat{\mathcal{M}}}^{\pi} \widehat{Q}^{k}-\lambda \eta(s, a) \\
    \leq& \mathcal{T}_{\mathcal{M}}^{\pi} \widehat{Q}^{k}+(1-f) \underbrace{\frac{C_{r, T, \delta} R_{\max }}{(1-\gamma) \sqrt{|\mathcal{D}(s, a)|}}}_{:=\Delta_{1}(s, a)}-\lambda \eta(s, a) +f(\underbrace{\left|r_{\mathcal{M}}-r_{\widehat{\mathcal{M}}}\right|+D\left(T_{\mathcal{M}}, T_{\widehat{\mathcal{M}}}\right) \frac{2 \gamma R_{\max }}{1-\gamma}}_{:=\Delta_{2}(s, a)}) \\
    =&\mathcal{T}^{\pi} \widehat{Q}^{k}+\underbrace{(1-f) \Delta_{1}(s, a)+f \Delta_{2}(s, a)-\lambda \eta(s, a)}_{:=\Delta(s, a)}
\end{aligned}
\end{equation}

\tx{By applying the fixed-point theorem, the above equation yields the following result:}
\begin{equation}\label{p-eq6}
\begin{aligned}
    \widehat{Q}^{\pi} &\leq \mathcal{T}^{\pi} \widehat{Q}^{\pi}+\Delta=r+\gamma P^{\pi} \widehat{Q}^{\pi}+\Delta \\
    \Rightarrow \widehat{Q}^{\pi} &\leq\left(I-\gamma P^{\pi}\right)^{-1}(r+\Delta)=\left(I-\gamma P^{\pi}\right)^{-1} r+\left(I-\gamma P^{\pi}\right)^{-1} \Delta
\end{aligned}
\end{equation}

Since $\mathcal{T}^{\pi} Q:=Q=r+\gamma P^{\pi} Q \Rightarrow Q=\left(I-\gamma P^{\pi}\right)^{-1} r$, the above equation can be further derived as:
\begin{equation}\label{p-eq7}
\widehat{Q}^{\pi}(s, a) \leq Q^{\pi}(s, a)+\left(I-\gamma P^{\pi}\right)^{-1} \Delta(s, a)
\end{equation}

Based on the value function $V^{\pi}(s)=\mathbb{E}_{a \sim \pi(a \mid s)}\left[Q^{\pi}(s, a)\right]$, it can derive:
\begin{equation}\label{p-eq8}
\widehat{V}^{\pi}(s) \leq V^{\pi}(s)+\left(I-\gamma P^{\pi}\right)^{-1} \mathbb{E}_{a \sim \pi(a \mid s)}[\Delta(s, a)]
\end{equation}
where $V^{\pi}(s)$ is the true value and $\widehat{V}^{\pi}(s)$ is the estimated value (the low bound). The lower bound value depends on the term $\left(I-\gamma P^{\pi}\right)^{-1} \mathbb{E}_{a \sim \pi(a \mid s)}\left[\Delta(s, a)\right]$, which also measures the gap between the true value and the lower bound. This gap is influenced by sampling error $\Delta_1(s,a)$, environment model error $\Delta_2(s,a)$ and the degree to which the data is far away from the offline region $\eta(s,a)$.

Here, to ensure the lower bound, $\lambda$ can be chosen to avoid overestimation. Moreover, since the expression $\left(I-\gamma P^{\pi}\right)^{-1}$ is positive semi-definite, as long as $\mathbb{E}_{a \sim \pi(a \mid s)}[\Delta(s, a)]<0$, the choice of $\lambda$ can be controlled by the following condition:
\begin{equation}\label{p-eq9}
\begin{aligned}
    \lambda \cdot \min _{s} \mathbb{E}_{a \sim \pi(a \mid s)}[\eta(s, a)] \geq \max _{s} \mathbb{E}_{a \sim \pi(a \mid s)}\left[(1-f) \Delta_{1}(s, a)+f \Delta_{2}(s, a)\right]
\end{aligned}
\end{equation}

\tx{Since $\Delta_{1}(s, a)>0, \Delta_{2}(s, a)>0$ and $\lambda>0$, for Eq. \eqref{p-eq9} to satisfy the given condition, it is necessary that the term $\min \mathbb{E}_{a \sim \pi(a \mid s)}[\eta(s, a)]$ is strictly positive, which leads to:}
\begin{equation}\label{p-eq10}
\begin{aligned}
    \mathbb{E}_{a \sim \pi(a \mid s)}[\eta(s, a)]=&\mathbb{E}_{a \sim \pi(a \mid s)}\left[\frac{\omega(s, a)-d(s, a)}{(1-f) d^{\pi_{b}}(s, a)+f d_{\widehat{\mathcal{M}}}^{\pi}(s, a)}\right]=\sum_a \frac{\omega(s, a)-d(s, a)}{(1-f) d^{\pi_{b}}(s) \frac{\pi_{b}(a \mid s)}{\pi(a \mid s)}+f d_{\mathcal{\widehat{\mathcal { M }}}}^{\pi}(s)}\\
    \geq& \frac{\sum_a \omega(s, a)}{(1-f) d^{\pi_{b}}(s) \max _{a}\left\{\frac{\pi_{b}(a \mid s)}{\pi(a \mid s)}\right\}+f d_{\widehat{\mathcal{M}}}^{\pi}(s)}-\frac{\sum_a d(s, a)}{(1-f) d^{\pi_{b}}(s) \min _{a}\left\{\frac{\pi_{b}(a \mid s)}{\pi(a \mid s)}\right\}+f d_{\widehat{\mathcal{\mathcal { M }}}}^{\pi}(s)} \\
\end{aligned}
\end{equation}

Consequently, the condition for $\min \mathbb{E}_{a \sim \pi(a \mid s)}[\eta(s, a)]>0$ to hold is derived as follows:
\begin{equation}\label{p-eq11}
\begin{aligned}
    \sum_a \omega(s, a)>\frac{(1-f) d^{\pi_{b}}(s) \max _{a}\left\{\frac{\pi_{b}(a \mid s)}{\pi(a \mid s)}\right\}+f d_{\widehat{\mathcal{M}}}^{\pi}(s)}{(1-f) d^{\pi_{b}}(s) \min _{a}\left\{\frac{\pi_{b}(a \mid s)}{\pi(a \mid s)}\right\}+f d_{\widehat{\mathcal{M}}}^{\pi}(s)} \sum_a d(s, a)=\xi(s)\sum_a d(s, a)
\end{aligned}
\end{equation}


\tx{By substituting Eq. \eqref{p-eq11} into \eqref{p-eq9}, we can further derive the following:}
\begin{equation}\label{p-eq12}
\begin{aligned}
    &\lambda \cdot \min _{s} \frac{(\xi(s)-1) \sum_a d(s, a) }{(1-f) d^{\pi_{b}}(s) \max _{a}\left\{\frac{\pi_{b}(a \mid s)}{\pi(a \mid s)}\right\}+f d_{\widehat{\mathcal{M}}}^{\pi}(s)} \\
    \geq&(1-f) \frac{C_{r, T, \delta} R_{\max }}{(1-\gamma) \min _{s, a}\{\sqrt{|\mathcal{D}(s, a)|}\}} +f\left[\max _{s, a}\left\{\left|r_{\mathcal{M}}-r_{\widehat{\mathcal{M}}}\right|\right\}+\frac{2 \gamma R_{\max }}{1-\gamma} \max _{s, a}\left\{D\left(T_{\widehat{\mathcal{M}}}, T_{\mathcal{M}}\right)\right\}\right]\\		
    \Rightarrow &\lambda \geq\frac{(1-f) \frac{C_{r, T, \delta} R_{\max }}{(1-\gamma) \min _{s, a}\{\sqrt{|\mathcal{D}|}\}}}{\min _{s}\left\{\frac{(\xi(s)-1) \sum_a d(s, a) } {\left[(1-f) d^{\pi_{b}}(s) \max _{a}\left\{\frac{\pi_{b}(a \mid s)}{\pi(a \mid s)}\right\}+f d_{\widehat{\mathcal{M}}}^{\pi}(s)\right]}\right\}}+\frac{f[\max \limits_{s, a}\left\{\left|r_{\mathcal{M}}-r_{\mathcal{M}}\right|\right\}+\frac{2 \gamma R_{\max }}{1-\gamma} \max\limits _{s, a}\left\{D_{T V}\left(T_{\mathcal{M}}, T_{\widehat{\mathcal{M}}}\right)\right\}]}
    {\min _{s}\left\{\frac{(\xi(s)-1) \sum_a d(s, a)} {\left[(1-f) d^{\pi_{b}}(s) \max _{a}\left\{\frac{\pi_{b}(a \mid s)}{\pi(a \mid s)}\right\}+f d_{\widehat{\mathcal{M}}}^{\pi}(s)\right]}\right\}}
\end{aligned}
\end{equation}

\tx{This completes the proof of Theorem \ref{thm1}.}

\subsection{Proof of Theorem \ref{thm2}}\label{proof2}
\tx{Following the previous work \cite{ref17}, they give the expression: $\widehat{Q}^{k+1}(s,a)=(\widehat{\mathcal{T}}^{\pi} \widehat{Q}^{k})(s, a)-\lambda[\frac{\rho(s, a)-d(s, a)}{(1-f) d(s, a)+f \rho(s, a)}]$. Assuming DOMAIN and COMBO methods have the same sampling and model errors, we have:}
\begin{equation}\label{p-eq13}
\begin{aligned}
    \kappa_{1}-\kappa_{2} =&\mathbb{E}_{s \sim d(s), a \sim \pi(a \mid s)}\left[\widehat{Q}_{1}^{\pi}(s, a)-\widehat{Q}_{2}^{\pi}(s, a)\right] \\
    =&\mathbb{E}_{s \sim d(s)}\left[\sum_a \frac{\pi(a \mid s)\left[\rho(s, a)-\omega(s, a)\right]}{(1-f) d^{\pi_{b}}(s) \pi_{b}(a \mid s)+f d_{\widehat{\mathcal{M}}}^{\pi}(s) \pi(a \mid s)} \right] \\
    \geq &\mathbb{E}_{s \sim d(s)}\left[\frac{\sum_a(\rho(s, a)-\omega(s, a))}{(1-f) d^{\pi_{b}}(s) \max _{a}\left\{\frac{\pi_{b}(a \mid s)}{\pi(a \mid s})\right\}+f d_{\widehat{\mathcal{M}}}^{\pi}(s)}\right] \\
    \geq &\underbrace{\frac{\mathbb{E}_{s \sim d(s)}\left[d_{\widehat{\mathcal{M}}}^{\pi}(s)\right]-\mathbb{E}_{s \sim d(s), a \sim \pi(a \mid s)}[\omega(s, a)]}{\max _{s}\left\{(1-f) d^{\pi_{b}}(s) \max _{a}\left\{\frac{\pi_{b}(a \mid s) }{ \pi(a \mid s)}\right\}+f d_{\widehat{\mathcal{M}}}^{\pi}(s)\right\}}}_{>0}
\end{aligned}
\end{equation}
where $d(s)$ represents the state distribution of the dataset, which can be an offline dataset, a model dataset, or the overall dataset. When $\mathbb{E}_{s \sim d(s)}[d_{\widehat{\mathcal{M}}}^{\pi}(s)] \geq \mathbb{E}_{s \sim d(s), a \sim \pi(a \mid s)}[\omega(s, a)]$, the average value function of the DOMAIN method on the dataset is greater than that of COMBO. In other words, under the aforementioned condition, DOMAIN is less conservative than COMBO. Thus, Theorem \ref{thm2} is proven.

\subsection{Proof of Theorem \ref{thm3}}\label{proof3}
RL aims to maximize the cumulative return, denoted as $\pi^{*}=\max _{\pi} J(\mathcal{M}, \pi):=\frac{1}{1-\gamma} \mathbb{E}_{(s, a) \sim d_{\mathcal{M}}^{\pi}(s, a)}[r(s, a)]$. Before proving Theorem \ref{thm3}, we present two lemmas.
\begin{lemma}\label{le2}
For any $\mathcal{M}$ and the experience $\overline{\mathcal{M}}$ generated by sampling according to the behavioral policy $\pi_{b}$, the performance of both $\mathcal{M}$ and $\overline{\mathcal{M}}$ under policy $\pi$ satisfies:
\begin{equation}\label{p-eq14}
    \begin{aligned}
        |J(\mathcal{M}, \pi)-J(\overline{\mathcal{M}}, \pi)| \leq\frac{1}{1-\gamma} \mathbb{E}_{(s, a) \sim d_{\overline{\mathcal{M}}}^{\pi}(s, a)}\left[\frac{C_{r, \delta}+R_{\max } C_{T, \delta}}{\sqrt{|\mathcal{D}(s, a)|}}\right]
    \end{aligned}
\end{equation}
\end{lemma}

\tx{\textbf{Proof for Lemma \ref{le2}.}} \tx{The proof follows a methodology similar to that of Kumar et al. \cite{ref10}. First, we apply the triangle inequality to decouple the reward return from the dynamic properties:}
\begin{equation}\label{p-eq15}
\begin{aligned}
    &|J(\mathcal{M}, \pi)-J(\overline{\mathcal{M}}, \pi)|\\
    =&\frac{1}{1-\gamma}\left|\sum_{s,a} \Big[d_{\overline{\mathcal{M}}}^{\pi}(s) \pi(a|s) r_{\overline{\mathcal{M}}}-d_{\mathcal{M}}^{\pi}(s) \pi(a|s) r_{\mathcal{M}}\Big]\right| \\
    \leq & \frac{1}{1-\gamma}\left|\sum_{s,a} d_{\overline{\mathcal{M}}}^{\pi}(s) \pi(a|s)\big[r_{\overline{\mathcal{M}}}(s, a)-r_{\mathcal{M}}(s, a)\big]\right|+\frac{1}{1-\gamma}\left|\sum_{s,a}\left(d_{\overline{\mathcal{M}}}^{\pi}(s)-d_{\mathcal{M}}^{\pi}(s)\right) \pi(a|s) r_{\mathcal{M}}(s, a)\right|
\end{aligned}
\end{equation}

Based on Assumption \ref{as1}, it can express the first term above as:
\begin{equation}\label{p-eq16}
\begin{aligned}
    \left|\sum_{s,a} d_{\overline{\mathcal{M}}}^{\pi} \pi\left[r_{\overline{\mathcal{M}}}-r_{\mathcal{M}}\right]\right|\leq \mathbb{E}_{(s, a) \sim d_{\overline{\mathcal{M}}}^{\pi}}\left[\frac{C_{r, \delta}}{\sqrt{|\mathcal{D}(s, a)|}}\right]
\end{aligned}
\end{equation}

Following the proof of Kumar \emph{et al}. \cite{ref10} on the upper bound of $\left|d_{\overline{\mathcal{M}}}^{\pi}(s)-d_{\mathcal{\mathcal { M }}}^{\pi}(s)\right|$, we have:
\begin{equation}\label{p-eq17}
\begin{aligned}
    \left|d_{\overline{\mathcal{M}}}^{\pi}(s)-d_{\mathcal{M}}^{\pi}(s)\right| \leq &\sum_{s,s'} \left|\int_{\mathcal{A}}\left(T_{\overline{\mathcal{M}}}-T_{\mathcal{M}}\right) \pi \right| d_{\overline{\mathcal{M}}}^{\pi}(s)
\end{aligned}
\end{equation}

From Assumption \ref{as1}, it can be deduced that
\begin{equation}
    \left|d_{\overline{\mathcal{M}}}^{\pi}(s)-d_{\mathcal{M}}^{\pi}(s)\right| \leq \mathbb{E}_{(s, a) \sim d_{\mathcal{M}}^{\pi}(s, a)}\left[\frac{C_{T, \delta}}{\sqrt{|\mathcal{D}(s, a)|}}\right]
\end{equation}

\tx{This result establishes an upper bound for the second term in Eq. \eqref{p-eq15}:}
\begin{equation}\label{p-eq18}
\begin{aligned}
    \left|\sum_{s,a}\left[d_{\overline{\mathcal{M}}}^{\pi}(s)-d_{\mathcal{M}}^{\pi}(s)\right] \pi(a|s) r_{\mathcal{M}}(s, a)\right| \leq \mathbb{E}_{(s, a) \sim d_{\mathcal{\overline{\mathcal { M }}}}^{\pi}(s, a)}\left[\frac{R_{\max } C_{T, \delta}}{\sqrt{|\mathcal{D}(s, a)|}}\right]
\end{aligned}
\end{equation}

Further derivation leads to:
\begin{equation}\label{p-eq19}
\begin{aligned}
    |J(\mathcal{M}, \pi)-J(\overline{\mathcal{M}}, \pi)| \leq \frac{1}{1-\gamma} \mathbb{E}_{(s, a) \sim d_{\overline{\mathcal{M}}}^{\pi}(s, a)}\left[\frac{C_{r, \delta}+R_{\max } C_{T, \delta}}{\sqrt{|\mathcal{D}(s, a)|}}\right]
\end{aligned}
\end{equation}

Lemma \ref{le2} is thus proven.

\begin{lemma}\label{le3}
For any $\mathcal{M}$ and an MDP $\widehat{\mathcal{M}}$ learned on the offline dataset, their performance on policy $\pi$ satisfies:
\begin{equation}\label{p-eq20}
    \begin{aligned}
        &|J(\mathcal{M}, \pi)-J(\widehat{\mathcal{M}}, \pi)| \leq \frac{1}{1-\gamma} \mathbb{E}_{(s, a) \sim d_{\widehat{\mathcal{M}}}^{\pi}(s, a)}\left[\varepsilon_{r}(s, a)+R_{\max } D(s, a)\right]
    \end{aligned}
\end{equation}
\end{lemma}

\tx{\textbf{Proof for Lemma \ref{le3}.}} The proof follows a similar process as in Lemma \ref{le2}. First, it can deduce:
\begin{equation}\label{p-eq21}
\begin{aligned}
    & |J(\mathcal{M}, \pi)-J(\widehat{\mathcal{M}}, \pi)|\\
    \leq & \frac{1}{1-\gamma}\left|\sum_{s,a} d_{\widehat{\mathcal{M}}}^{\pi}(s) \pi(a \mid s)\left[r_{\widehat{\mathcal{M}}}(s, a)-r_{\mathcal{M}}\right]\right|+\frac{1}{1-\gamma}\left|\sum_{s,a}\left(d_{\widehat{\mathcal{M}}}^{\pi}(s)-d_{\mathcal{M}}^{\pi}(s)\right) \pi(a|s) r_{\mathcal{M}}\right|
\end{aligned}
\end{equation}

Let $\left|r_{\widehat{\mathcal{M}}}(s, a)-r_{\mathcal{M}}(s, a)\right|=\varepsilon_{r}(s, a)$ denote the reward prediction error, and $\left|\left(T_{\overline{\mathcal{M}}}\left(s^{\prime} \mid s, a\right)-T_{\mathcal{M}}\left(s^{\prime} \mid s, a\right)\right)\right|_{1}=D(s, a)$ denote the state transition prediction error. Consequently, it can obtain:
\begin{equation}\label{p-eq22}
\begin{aligned}
    &|J(\mathcal{M}, \pi)-J(\widehat{\mathcal{M}}, \pi)| \leq \frac{1}{1-\gamma} \mathbb{E}_{(s, a) \sim d_{\widehat{\mathcal{M}}}^{\pi}(s, a)}\left[\varepsilon_{r}(s, a)+R_{\max } D(s, a)\right]
\end{aligned}
\end{equation}

Lemma \ref{le3} is thus proven. 

\tx{\textbf{Proof for Theorem \ref{thm3}.}} Let $\widehat{Q}^{\pi}(s, a)$ denote the fixed point of Eq. \eqref{eq14}. The policy update for the DOMAIN method is expressed as follows:
\begin{equation}\label{p-eq23}
\pi^{*}=\max _{\pi}\left[J\left(\mathcal{M}_{f}, \pi\right)-\frac{\lambda}{1-\gamma} \mathbb{E}_{(s, a) \sim d_{\mathcal{M}_{f}}^{\pi}}[\eta(s, a)]\right]
\end{equation}
where $J\left(\mathcal{M}_{f}, \pi\right)=(1-f) J(\overline{\mathcal{M}}, \pi)+f J(\widehat{\mathcal{M}}, \pi)$, $\mathcal{M}_{f}:=(1-f) \overline{\mathcal{M}}+f \widehat{\mathcal{M}}$, and $\eta(s, a)=(\omega(s, a)-d(s, a))/d_{\beta}(s, a)$. Since $\pi^{*}$ represents the optimal value in Eq. \eqref{p-eq23}, and let $\varpi(\pi, f)=\mathbb{E}_{(s, a) \sim d_{\mathcal{M}_{f}}^{\pi}}[\eta(s, a)]$, we have:
\begin{equation}\label{p-eq24}
J\left(\mathcal{M}_{f}, \pi^{*}\right)-\frac{\lambda\varpi\left(\pi^{*}, f\right)}{1-\gamma}  \geq J\left(\mathcal{M}_{f}, \pi_{b}\right)-\frac{\lambda\varpi\left(\pi_{b}, f\right)}{1-\gamma} 
\end{equation}

In order to compare the performance of the learned policy $\pi^{*}$ and the behavioral policy $\pi_{b}$ on the actual MDP, Theorem \ref{thm3} employs the triangle inequality to deduce:
\begin{equation}\label{p-eq25}
\begin{aligned}
     \left|J\left(\mathcal{M}_{f}, \pi\right)-J(\mathcal{M}, \pi)\right| = &|(1-f) J(\overline{\mathcal{M}}, \pi)+f J(\widehat{\mathcal{M}}, \pi)-J(\mathcal{M}, \pi)| \\
    \leq & (1-f)|J(\mathcal{M}, \pi)-J(\overline{\mathcal{M}}, \pi)|+f|J(\mathcal{M}, \pi)-J(\widehat{\mathcal{M}}, \pi)|
\end{aligned}
\end{equation}

By utilizing Lemma \ref{le2}-\ref{le3}, it can infer:
\begin{equation}\label{p-eq26}
\begin{aligned}
    &\left|J\left(\mathcal{M}_{f}, \pi\right)-J(\mathcal{M}, \pi)\right| \\
    \leq&(1-f)|J(\mathcal{M}, \pi)-J(\overline{\mathcal{M}}, \pi)|+f|J(\mathcal{M}, \pi)-J(\widehat{\mathcal{M}}, \pi)| \\
    \leq& \frac{1-f}{1-\gamma} \mathbb{E}_{(s, a) \sim d_{\mathcal{M}}^{\pi}(s, a)}\left[\frac{C_{r, \delta}+R_{\max } C_{T, \delta}}{\sqrt{|\mathcal{D}(s, a)|}}\right] +\frac{f}{1-\gamma} \mathbb{E}_{(s, a) \sim d_{\mathcal{M}}^{\pi}(s, a)}\left[\varepsilon_{r}(s, a)+R_{\max } D(s, a)\right]
\end{aligned}
\end{equation}

Substituting Eq. \eqref{p-eq26} into \eqref{p-eq24}, it can further derive:
\begin{equation}\label{p-eq27}
\begin{aligned}
    &J\left(\mathcal{M}, \pi^{*}\right)-J\left(\mathcal{M}, \pi_{b}\right) \\
    \geq& \frac{\lambda}{1-\gamma}\left[\varpi\left(\pi^{*}, f\right)-\varpi\left(\pi_{b}, f\right)\right] -\frac{1-f}{1-\gamma}\left\{\mathbb{E}_{(s, a) \sim d_{\mathcal{M}}^{\pi^*}}\left[\frac{C_{r, \delta}+R_{\max } C_{T, \delta}}{\sqrt{|\mathcal{D}(s, a)|}}\right]\right.\left.+\mathbb{E}_{(s, a) \sim d_{\mathcal{M}}^{\pi_{b}}}\left[\frac{C_{r, \delta}+R_{\max } C_{T, \delta}}{\sqrt{|\mathcal{D}(s, a)|}}\right]\right\} \\
    &-\frac{f}{1-\gamma}\left\{\mathbb{E}_{(s, a) \sim d_{\mathcal{M}}^{\pi^{*}}}\left[\varepsilon_{r}(s, a)+R_{\max } D(s, a)\right]\right.\left.+\mathbb{E}_{(s, a) \sim d_{\mathcal{M}}^{\pi_{b}}}\left[\varepsilon_{r}(s, a)+R_{\max } D(s, a)\right]\right\}
\end{aligned}
\end{equation}

\tx{The behavioral policy $\pi_{b}$ is derived from the offline dataset $\mathcal{D}_{\text {offline }}$, resulting in a larger $\sqrt{|\mathcal{D}(s, a)|}$ under the distribution $d_{\overline{\mathcal{M}}}^{\pi_{b}}$, which indicates smaller sampling errors. Furthermore, since the environment model is trained using $\mathcal{D}_{\text {offline }}$, the model error is reduced under the distribution $d_{\widehat{\mathcal{M}}}^{\pi_{b}}$. Therefore, Eq. \eqref{p-eq27} can be further simplified as:}
\begin{equation}\label{p-eq28}
\begin{aligned}
    & J\left(\mathcal{M}, \pi^{*}\right)-J\left(\mathcal{M}, \pi_{b}\right) \\
    \geq &\frac{\lambda}{1-\gamma}\left[\varpi\left(\pi^{*}, f\right)-\varpi\left(\pi_{b}, f\right)\right]-2 \frac{1-f}{1-\gamma} \mathbb{E}_{(s, a) \sim d_{\mathcal{M}}^{\pi^{*}}}\left[\frac{C_{r, \delta}+R_{\max } C_{T, \delta}}{\sqrt{|\mathcal{D}(s, a)|}}\right] -2 \frac{f}{1-\gamma} \mathbb{E}_{(s, a) \sim d_{\mathcal{M}}^{\pi^{*}}}\Big[\varepsilon_{r}(s, a)+R_{\max } D(s, a)\Big]
\end{aligned}
\end{equation}

\tx{This completes the proof of Theorem \ref{thm3}.}